\documentclass[letterpaper]{article} 
\usepackage[preprint]{aaai2027}
\usepackage[hyphens]{url}  
\usepackage{graphicx} 
\usepackage{natbib}  
\usepackage{caption} 
\usepackage{algorithm}
\usepackage{algorithmic}
\usepackage{amssymb}
\usepackage{newfloat}
\usepackage{tabularx}
\usepackage{array}
\usepackage{listings}
\DeclareCaptionStyle{ruled}{labelfont=normalfont,labelsep=colon,strut=off} 
\floatstyle{ruled}
\newfloat{listing}{tb}{lst}{}
\floatname{listing}{Listing}

\usepackage{booktabs}
\usepackage{xcolor}
\usepackage{amsmath}
\newcommand{\method}{OD-Gear}
\newcommand{\clip}{\operatorname{clip}}
\DeclareMathSizes{10}{9}{7}{5}
\usepackage{siunitx}
\newenvironment{inlinetable}
  {\par\begingroup\setlength{\topsep}{3pt}\setlength{\partopsep}{0pt}\setlength{\abovecaptionskip}{5pt}\begin{center}}
  {\end{center}\endgroup}

\title{OD-Gear: Online Decomposition and Group Sampling for Expert-Guided Adversarial Routing in Scalable Capacitated Vehicle Routing}
\author{
    Dongbin Jiao\textsuperscript{\rm 1},
    Zisheng Chen\textsuperscript{\rm 1},
    Xianyi Wang\textsuperscript{\rm 1},
    Jintao Shi\textsuperscript{\rm 1},\\
    Shengcai Liu\textsuperscript{\rm 2},
    Shi Yan\textsuperscript{\rm 1}
}
\affiliations{}

\begin{document}

\maketitle
\begingroup
\renewcommand{\thefootnote}{}
\footnotetext{\raggedright\textsuperscript{\rm 1}School of Information Science and Engineering, Lanzhou University, Lanzhou 730000, P. R. China (Emails: \{jiaodb, chzisheng2024, wxianyi2025, shijt2024\}@lzu.edu.cn). \textsuperscript{\rm 2}Guangdong Provincial Key Laboratory of Brain-Inspired Intelligent Computation, Department of CSE, SUSTech, Shenzhen 518055, P. R. China. Correspondence to: Shengcai Liu $<$liusc3@sustech.edu.cn$>$, Shi Yan $<$yanshi@lzu.edu.cn$>$.}
\endgroup

\begin{abstract}
Solving large-scale capacitated vehicle routing problems (CVRP) is hindered by the high complexity of classical heuristics and the limited generalization of neural solvers. To bridge this gap, we propose OD-Gear, an expert-guided adversarial framework that integrates hybrid genetic search (HGS) and online barycenter clustering (BCC) decomposition with group-relative optimization. 
OD-Gear internalizes expert heuristics into a graph attention network (GAT)-based policy via high-fidelity knowledge distillation. Our minimax adversarial training distills divide-and-conquer strategies into dense surrogate rewards, while a group-sampling strategy  exploits relative solution advantages to promote both diversity and quality. This architecture enables high-quality, clustering-free inference on massive graphs, effectively bypassing the overhead of traditional decomposition. Empirical results demonstrate that OD-Gear achieves state-of-the-art (SOTA) performance across most benchmarks, remaining highly competitive at the 10,000-node scale. By providing heuristic-quality solutions with low-latency, OD-Gear offers a robust and scalable framework for large-scale CVRP.


\end{abstract}


\section{Introduction}
The capacitated vehicle routing problem (CVRP) is a fundamental NP-hard challenge 
in modern logistics and supply chain management \cite{toth2014vehicle,laporte2024fifty}. Due to the combinatorial explosion of the search space, traditional exact or heuristic solvers incur  prohibitive computational costs on large-scale instances \cite{bogyrbayeva2024machine}. 

Neural combinatorial optimization (NCO) has emerged as a 
transformative paradigm for approximate routing policies via end-to-end inference \cite{wu2026neural}. However, current neural solvers face a critical scalability-optimality bottleneck: supervised approaches are often restricted by static, expensive labels, while reinforcement learning (RL) methods typically suffer from sparse reward signals and poor generalization \cite{bello2017neural,gunarathna2022dynamic}. Furthermore, standard Transformer-based architectures frequently are prone to mode collapse, which limits solution diversity and structural integrity on massive graphs.

To address these limitations, generative flow networks (GFlowNets)  \cite{bengio2021flow,zhang2023let} have been proposed to sample diverse trajectories with probabilities proportional to a reward function. 
Recent flow-based models like AGFN \cite{zhang2025adversarial} improves solution diversity but neglect local signals, failing to capture the fine-grained structures necessary for large-scale optimization. Similarly, while HBG \cite{NEURIPS2025_1898054a} unifies trajectory and detailed balance, it remains constrained by its reliance on base GFlowNet solvers. Ultimately, both frameworks lack the structural inductive biases and expert regularization essential for robust, large-scale performance.


In this work, we propose OD-Gear, an expert-guided adversarial routing framework with online decomposition and group sampling.  Grounded in the GFlowNet paradigm, OD-Gear bridges the gap between the inference efficiency of neural methods and the solution quality of state-of-the-art (SOTA) heuristics. Unlike standard RL, OD-Gear reformulates route construction as a stochastic flow matching problem. By utilizing the trajectory balance (TB) objective, the model learns a policy that ensures precise distributional alignment with a high-performance expert manifold.

OD-Gear utilizes a graph attention network (GAT) generative policy designed to capture intricate dependencies between spatial proximity, customer demands, and vehicle capacity. To overcome reward sparsity, we introduce a dynamic expert oracle that synergizes hybrid genetic 
search (HGS) \cite{vidal2012hybrid} with online barycenter clustering (BCC) decomposition \cite{santini2023decomposition} . Through a minimax adversarial framework, this oracle provides high-fidelity ``divide-and-conquer'' demonstrations, distilled into dense surrogate rewards. Furthermore, group sampling exploits relative solution advantages to enhance both diversity and quality. By internalizing these partitioning and refinement strategies, the generator achieves high-quality, clustering-free inference. Empirical results show that OD-Gear achieves competitive real-time CVRP performance through near-constant per-node inference latency to large-scale instances.


Our contributions are as follows:
(i) We propose OD-Gear, an adversarial GFlowNet framework integrating online BCC decomposition with group-relative optimization to bridge neural inference and classical heuristics. 
(ii) We introduce a hybrid expert oracle (HGS + BCC) that distills complex decomposition strategies into dense surrogate rewards via minimax training, enabling high-quality, clustering-free inference. (iii) OD-Gear achieves SOTA performance across most benchmarks and maintains high efficiency at the 10,000-node scale, effectively overcoming the scalability bottlenecks of large-scale CVRP.



\section{Preliminaries and Related Work}
The CVRP is generally defined on a complete graph $G(V, E)$ with a depot $v_0$ and $N$ customer nodes $V_c = \{v_1,\dots,v_N\}$. Each edge $(i, j) \in E$ has 
a non-negative travel cost $c_{ij}$, and each node $i \in V_c$ has a demand $d_i$. We aim to minimize $\sum c_{ij}$ for $K$ vehicles of capacity $Q$, such that each customer is visited exactly once, route demands do not exceed $Q$,
and all routes start and end at $v_0$.

\textbf{Heuristics  and Decomposition for Large-scale VRP.} 
The SOTA VRP 
performance relies on heuristics like HGS \cite{vidal2012hybrid,vidal2022hybrid} and LKH-3 \cite{helsgaun2017extension}. These solvers achieve near-optimal accuracy through population-based diversity management and sophisticated local search. However, they encounter a "scalability wall" as the combinatorial search space expands, rendering global optimization on massive instances computationally prohibitive. To mitigate this, research has shifted toward decomposition strategies \cite{zong2022rbg,santini2023decomposition,ye2024glop,zheng2024udc,kerscher2025decompose}, specifically clustering-based approaches that partition global problems into tractable sub-problems. OD-Gear leverages these high-performance heuristics not as standalone solvers, but as a dynamic expert oracle to provide high-fidelity supervisory signals for neural training.

\textbf{Neural Combinatorial Optimization (NCO).}  NCO typically models the CVRP as an autoregressive process, optimizing  $P_{\theta}(\boldsymbol{a}|s) = \prod_{t=1}^{T} p_{\theta}(a_t | s, \boldsymbol{a}_{<t})$ to maximize expected rewards. NCO reframes the VRP as a data-driven policy approximation problem \cite{bengio2021machine,zheng2024dpn}. Constructive neural solvers typically utilize graph neural networks (GNNs) or Transformers to generate solutions end-to-end \cite{kool2019attention,kotary2021end,cappart2023combinatorial,luo2025boosting}. While architectures like GAT effectively capture heterogeneous node features and spatial dependencies \cite{bui2023imitation,cappart2023combinatorial,luo2023neural}, most solvers rely on policy gradient methods (e.g., REINFORCE \cite{williams1992simple}) that struggle with sparse reward signals and poor generalization to large-scale distributions \cite{bengio2021machine,yao2025rethinking,wu2026neural}. Furthermore, standard architectures often suffer from mode collapse, failing to represent the multimodal nature of optimal routing for large-scale CVRP. We address these limitations by adopting the GFlowNet paradigm \cite{bengio2023gflownet}, which facilitates robust exploration and better represents the solution distribution through flow consistency.

\textbf{Adversarial Learning and Knowledge Distillation (KD).}
KD via imitation learning is a common approach to bridge the gap between heuristic precision and neural speed \cite{bi2022learning,sun2023difusco,bogyrbayeva2024machine}. However, pure imitation often suffers from exposure bias and fails to generalize when the student deviates from expert trajectories. GFlowNets mitigate this by framing optimization as a distribution matching task, forcing the model to internalize the structural manifold of expert solutions \cite{kim2024genetic,kim2025ant,gui2026chainofcontext}. Recent advancements, such as AGFN \cite{zhang2025adversarial,NEURIPS2025_1898054a}, integrate adversarial objectives with flow-based sampling to provide denser training signals. OD-Gear integrates generative adversarial networks (GANs) and decomposition‑aware heuristics via adversarial learning with an embedded expert, internalizing ``divide-and-conquer'' logic for efficient, near‑optimal large‑scale inference.

\section{Methodology}
\begin{figure*}[htp]
    \centering    \includegraphics[width=\textwidth]{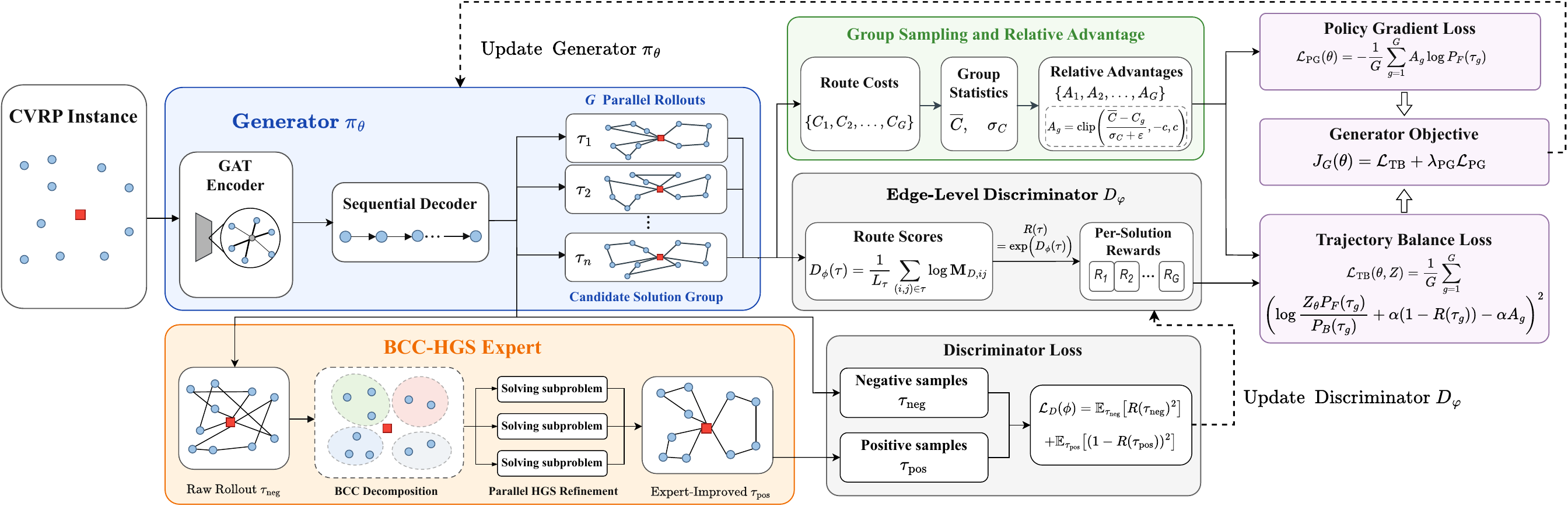}
    \caption{Overall architecture of the OD-Gear framework.} 
    \label{fig:schematic} 
\end{figure*}

As illustrated in Figure \ref{fig:schematic}, OD-Gear integrates neural inference with heuristic-driven decomposition via an adversarial loop. A GAT-based policy performs exploratory rollouts, which are refined by a BCC-HGS expert oracle to provide high-fidelity supervisory signals through an edge-level discriminator. By distilling these expert strategies into dense rewards and leveraging a group-sampling strategy to enhance solution diversity and quality, OD-Gear internalizes complex  ``divide-and-conquer” logic, facilitating  high-quality, clustering-free inference at runtime.
\subsection{GAT-Based Generator}
The GAT-based generator $\pi_\theta$ serves as 
OD-Gear's core decision-making component for autonomous inference. To capture intricate node interactions, we adopt a GAT backbone  \cite{velivckovic2018graph}, which uses multi-head self-attention to adaptively weight neighboring nodes based on their latent features. This mechanism is uniquely suited for the CVRP by effectively modeling the complex interdependencies among spatial, heterogeneous demands, and residual vehicle capacity. By effectively aggregating these  features, the GAT aggregates features for expressive representations critical to high-quality routing.  

\textbf{Structural Encoder}.
A multi-layer GAT encodes topological and attribute-based features of the CVRP instance into high-dimensional node embeddings vectors $\mathbf{h}_i \in \mathbb{R}^{D_{\text{units}}}$ by capturing intricate node dependencies through three stages:

\emph{Initial Feature Transformation.}
To harmonize input features of varying dimensions, we first apply independent linear projections to the raw node features $\mathbf{x}_i$ and edge features $\mathbf{a}_{ij}$. This maps them into a unified hidden dimension $D_{\text{units}}$, providing a standardized input for the attention mechanism:
\begin{align}
\mathbf{h}_i^{(0)} &= \sigma\!\left( \mathbf{W}_{\text{node}} \mathbf{x}_i + \mathbf{b}_{\text{node}} \right), \\
\mathbf{e_{ij}} &= \sigma\!\left( \mathbf{W}_{\text{edge}} \mathbf{a}_{ij} + \mathbf{b}_{\text{edge}} \right),
\end{align}
where $\mathbf{W}_{\text{node}}$ and $\mathbf{W}_{\text{edge}}$ are trainable weight matrices, $\mathbf{b}_{\text{node}}$ and $\mathbf{b}_{\text{edge}}$ are bias vectors, and $\sigma(\cdot)$ denotes a non-linear activation function.

\emph{Node Representation Update.} The encoder consists of $L$ stacked GAT layers. In each layer, the model performs message passing by computing attention scores based on the source node, its neighbors, and the associated edge features. For each attention head, the unnormalized correlation score $s_{ij}^{(k,h)}$ between node $i$
and $j$ is computed as
\begin{equation}
    s_{ij}^{(k,h)} = \text{AttentionMech}^{(k,h)}(\mathbf{h}_i^{(k)}, \mathbf{h}_j^{(k)}, \mathbf{e}_{ij}).
\end{equation}
To ensure a probabilistic interpretation and facilitate comparison across the neighborhood $\mathcal{N}(i)$, these scores are normalized using the softmax function to obtain the attention coefficients $\alpha_{ij}^{(k,h)}$:
\begin{equation}
\alpha_{ij}^{(k,h)} = \operatorname{softmax}_j\left(s_{ij}^{(k,h)}\right).
\end{equation}
These coefficients $\alpha_{ij}^{(k,h)}$ signify the relative importance of neighbor $j$ to node $i$ within the $h$-th subspace.

\emph{Multi-head Aggregation}. The encoder aggregates the neighborhood information via multi-head attention. To stabilize training  and mitigate vanishing gradients, we incorporate residual connections and batch normalization (BN):
\begin{equation}
\mathbf{h}_i^{(k+1)} = \mathbf{h}_i^{(k)} +
\mathrm{BN}^{(k)}\!\left( \sigma\!\left( \mathbf{h}_{\mathrm{aggr},i}^{(k+1)} \right) \right),
\end{equation}
where $\mathbf{h}_{\mathrm{aggr},i}^{(k+1)}$ represents the concatenated or averaged output from the multiple attention heads. After $L$ iterations, the final embedding $\mathbf{h}_i$ encapsulates the global topological structure of the graph, providing a context-rich representation for the subsequent sequential decision-making process.

\textbf{Sequential Constructive Decoder}.
OD-Gear employs an autoregressive decoder that parameterizes  a policy network  $\pi_{\theta}(a_t|s_t)$ to construct feasible routing  sequences. 

At each decoding step $t$ with current visited node $v_i$, the decoder evaluates transition propensities to candidate neighbor $v_j$ using the  GAT-encoded embeddings.
Specifically, we construct a relational context vector $f_{ij}$ by concatenating the source and target embeddings, which is then mapped to a scalar logit $u_{ij}$ via a multi-layer perceptron (MLP):
\begin{equation}
   \mathbf f_{ij} = [\mathbf h_i \, \| \, \mathbf h_j \, \| \, \mathbf e_{ij}], \quad u_{ij} = \text{MLP}_{\theta_{dec}}(\mathbf f_{ij}),
\end{equation}
where $[\cdot \| \cdot]$ represents the concatenation operation. This context vector $f_{ij}$ captures the heterogeneous dependencies between spatial proximity and demand-capacity interactions. To strictly enforce the hard constraints of the CVRP, we incorporate a dynamic masking mechanism. A mask $M_{ij}^{(t)}$ is applied to the raw logits, defined as
\begin{equation}
    M_{ij}^{(t)} = 
    \begin{cases} 
    0 & \text{if } v_j \in \mathcal{N}_{valid}(s_t) \\
    -\infty & \text{otherwise}
    \end{cases},
\end{equation}
where $\mathcal{N}_{valid}(s_t)$ represents the set of unvisited nodes that satisfy the capacity-feasible nodes. The selection probability follows a softmax distribution over the masked logits:
\begin{equation}
    p(a_t = v_j | s_t) = \frac{\exp(u_{ij} + M_{ij}^{(t)})}{\sum_{k \in \mathcal{P}} \exp(u_{ik} + M_{ik}^{(t)})}.
\label{eq:decoder}
\end{equation}
To balance between exploration and exploitation with the GFlowNet objective, we adopt a hybrid decoding mechanism \cite{zhang2025adversarial} .  
During training, we sample stochastically from the decoding distribution $p(\cdot|s_t)$ using a logarithmically increasing inverse temperature to transition from diverse exploration to exploitation.
At inference, we employ a $\rho$-greedy strategy that selects the greedy action with probability $1-\rho$ and samples stochastically otherwise. These stochastic steps diversify the parallel rollouts, allowing us to select the lowest-cost solution and effectively balance search-space exploration with high-quality construction.


\subsection{Decomposition-Augmented HGS Expert}
The expert policy $P_E$ acts as a high-fidelity supervisory oracle, distilling high-quality local and global routing logic into an upper-bound performance target  for the generator.
\begin{algorithm}[t]
\caption{Expert Solutions via BCC Decomposition}
\label{alg:expert}
\begin{algorithmic}[1]
\STATE \textbf{Input:} Initial solution $R_{\text{init}}$, node coordinates  $X$, demands $D$, capacity $Q$, target subproblem  size $m$.
\STATE \textbf{Output:} Refined global solution $R_{\text{final}}$, total cost $C_{\text{final}}$.

\STATE \textbf{// Phase 1: Route-Based Spatial Decomposition}
\STATE $N \leftarrow \text{CountCustomers}(R_{\text{init}})$
\STATE \emph{// Determine number of subproblems}
\STATE $k \leftarrow \lceil N / m \rceil$ 
\STATE $B \leftarrow$ ComputeBarycenters($R_{\text{init}}, X$)
\STATE $Labels \leftarrow$ KMeans($B, k$)
\STATE \emph{// Partition into independent clusters $\{C_1, \dots, C_k\}$}
\STATE $\mathcal{C} \leftarrow$ GroupRoutesByLabel($R_{\text{init}}, \text{Labels}$)

\STATE \textbf{// Phase 2: Parallel Subproblem Solving}
\STATE $\text{Solutions} \leftarrow \text{List of size } k$
\FORALL{cluster $C_{i}$ in $\mathcal{C}$, \textbf{in parallel}}
\STATE \text{//} Map global $v_{\text{id}}$ to local $[0, n')$
    \STATE $V_{\text{sub}}, \mathcal{M} \leftarrow$ BuildMapping($C_{i}, v_0$) 
    \STATE \emph{// Constraint: Vehicle count matches route count}
    \STATE $K_{\text{sub}} \leftarrow |C_{i}|$ 
    
    \STATE $Data_{\text{sp}} \leftarrow$ CreateInstance($V_{\text{sub}}, D, X, Q, K_{sub}$)
     \STATE \emph{// Solve subproblem}
    \STATE $R_{\text{local}} \leftarrow$ SolveHGS($Data_{\text{sp}}$) 
     \STATE \emph{// Store optimized routes with global IDs}
    \STATE $\text{Solutions}[i] \leftarrow$ MapToGlobal($R_{\text{local}}, \mathcal{M}$) 
\ENDFOR

\STATE \textbf{// Phase 3: Global Merge}
\STATE $R_{\text{final}} \leftarrow \emptyset$
\FORALL{$R_{\text{part}}$ in $\text{Solutions}$}
    \STATE $R_{\text{final}} \leftarrow R_{\text{final}} \cup R_{\text{part}}$ 
\ENDFOR
\STATE $C_{\text{final}} \leftarrow$ CalculateCost($R_{\text{final}}$)
\RETURN $R_{\text{\text{final}}}, C_{\text{\text{final}}}$.
\end{algorithmic}
\end{algorithm}

\textbf{Heuristic Selection.} We utilize the HGS \cite{vidal2012hybrid} as the expert algorithm. As a SOTA metaheuristic for the CVRP, HGS achieves  superior performance by synergizing the global exploration capabilities of genetic algorithms with the intensive exploitation of sophisticated local search.

\textbf{Large-Scale Decomposition Strategy.} 
To mitigate the performance degradation of standalone HGS on massive instances due to search space expansion
\cite{santini2023decomposition}, we incorporate a ``divide-and-conquer" framework based on BCC. As a route-based decomposition strategy, BCC offers high robustness by partitioning the problem into manageable sub-topologies while preserving the integrity of individual routes \cite{santini2023decomposition}.

\textbf{Expert Solution Generation.} 
As detailed in Algorithm \ref{alg:expert}, the generation of the expert solution $S_{\text{expert}}$ follows a three-stage process: (i) \emph{Decomposition}: Given an initial feasible solution $R_{\text{init}}$, we calculate route barycenters and apply $k$-means clustering to partition the global problem into $k$ geographically localized subproblems. (ii) \emph{Parallel optimization}: Each subproblem is optimized in parallel using HGS. This enables intensive, localized route refinement within each cluster, significantly reducing the computational burden compared to exhaustive global search.
(iii) \emph{Merge}: The optimized sub-solutions are re-integrated to form the final high-quality global solution $S_{\text{expert}}$.

Leveraging this decomposition-augmented oracle, the training process provides the neural policy with a distribution $P_E$ that reflects low-level routing logic and high-level topological partitioning essential for large-scale optimization.
\subsection{Adversarial Training}
To internalize expert strategies, OD-Gear employs a minimax adversarial game between the generator $\pi_{\theta}$ and a discriminator $D_{\phi}$, transforming sparse terminal rewards into dense feedback via distributional alignment. By minimizing the statistical divergence between the generated distribution $P_{\theta}(S)$ and the  expert distribution  $P_{E}(S)$ from HGS-BCC, the model learns to navigate the solution manifold defined by the expert  oracle depicted in Figure \ref{fig:schematic}. 

\textbf{Discriminator: Edge-Level Probabilistic Estimator}.
The discriminator $D_{\phi}$ acts as an edge-level probability estimator. While
sharing the GAT-based architecture with the generator, it serves a distinct evaluative role.
Given a CVRP graph $G = (\mathcal{V}, \mathcal{E})$, 
$D_{\phi}$ outputs an edge probability matrix $\mathbf{M}_D \in [0, 1]^{N \times N}$, where $\mathbf{M}_{D, ij}$ denotes the estimated likelihood that edge $(i, j)$ belongs to an optimal solution.
The discriminator evaluates a candidate solution $\tau = \{(v_1, v_2), \dots, (v_{k}, v_1)\}$ by aggregating the scores of its constituent edges. We calculate the length-normalized log-probability score $D_{\phi}(\tau)$ as
\begin{equation}
    D_{\phi}(\tau) = \frac{1}{L_{\tau}}\sum_{(i, j) \in \tau} \log \mathbf{M}_{D, ij} ,
\label{eq:reward}
\end{equation}
where $L_{\tau}$ denotes the total number of traversed edges 
in trajectory $\tau$. The dense surrogate reward $R(\tau) = \exp(D_{\phi}(\tau))$  reflects the joint confidence of all edges in the path.

\textbf{Optimization}.
The discriminator is optimized using a least squares GAN objective to distinguish between raw generated policies and their 
expert-improved counterparts. The training process involves two types of samples. Negative samples ($\tau_{neg}$) are trajectories drawn directly from the generator policy  $\pi_{\theta}$ without post-processing. The discriminator is optimized to minimize their reward, driving $R(\tau_{neg}) \to 0$. Positive samples ($\tau_{pos}$) are high-quality trajectories obtained by refining generator outputs with the expert policy augmented by BCC decomposition. These trajectories serve as approximations to the target distribution, and the discriminator is trained to maximize their reward, pushing $R(\tau_{pos}) \to 1$. The discriminator loss function is defined as
\begin{equation}
    \mathcal{L}_D(\phi) =\mathbb{E}_{\tau_{neg}} [R(\tau_{neg})^2] +\mathbb{E}_{\tau_{pos}} [(1 - R(\tau_{pos}))^2],
\label{eq:disc}
\end{equation}
where $\tau_{neg}$ are raw generator trajectories and $\tau_{pos}$ their expert-refined counterparts.

\subsection{Group Sampling Strategy}
\textbf{Relative Advantage Calculation}.
To mitigate the variance of single-trajectory estimates, eliminate the need for auxiliary value network, and foster solution diversity, we adopt a group-relative advantage formulation inspired by GRPO \cite{shao2024deepseekmath}. For each training instance, we sample $G$ trajectories $\{\tau_g\}_{g=1}^{G}\sim\pi_\theta$. Let $C_g$ be the total travel cost of $\tau_g$, $\bar C=\frac{1}{G}\sum_{g=1}^{G}C_g$ be the group mean cost,
and $\sigma_C$ be the population standard deviation of the sampled group. We define the group-relative advantage as
\begin{equation}
A_g=\clip\!\left(\frac{\bar C-C_g}{\sigma_C+\epsilon},-c,c\right),
\label{eq:advantage}
\end{equation}
where $\epsilon$ is a small constant and $c$ is the clip bound. Setting $A_g=0$ if $\sigma_C \leq \epsilon$, this mechanism assigns positive advantages to trajectories outperforming the group mean, effectively reinforcing superior route-finding behaviors.

\textbf{Generator Objective}.
We incorporate $A_g$ into the TB and auxiliary policy-gradient (PG) losses. The TB target is formulated as
\begin{equation}\label{eq:tb}
L_{TB}(\theta, Z) = \frac{1}{G} \sum_{g=1}^{G} \left[ \log \frac{Z_\theta P_F(\tau_g)}{P_B(\tau_g)} + \alpha (1 - R(\tau_g)) - \alpha A_g \right]^2,
\end{equation}
where $Z_\theta$ is the learnable partition function, $P_F(\tau_g)/P_B(\tau_g)$ denote the forward/backward probabilities, and $R(\tau_g)\in[0,1]$ is the discriminator reward of Eq.~\ref{eq:reward}, and $\alpha>0$ balances the adversarial penalty $1-R(\tau_g)$ and the advantage $A_g$. In experiments, $\alpha$ follows a logarithmic growth schedule (e.g., 500 to 2000 for CVRP-200), with endpoints scaling proportionally to the instance size. 
To further  guide the policy toward low-cost trajectories, we introduce an auxiliary PG term:
\begin{equation}\label{eq:pg}
\mathcal L_{\mathrm{PG}}=-\frac{1}{G}\sum_{g=1}^{G}A_g\log P_F(\tau_g),
\end{equation}
where $A_g$ serves as a constant baseline. The total generator loss  is 
\begin{equation}\label{eq:total}
J_G(\theta)=\mathcal L_{\mathrm{TB}}+\lambda_{\mathrm{PG}}\mathcal L_{\mathrm{PG}},
\end{equation}
where $\lambda_{\mathrm{PG}}$ is weighting factor, with advantages taken over the full group. Notably, OD-Gear operates without auxiliary value networks, PPO likelihood ratios, or KL-divergence regularizers.

\begin{algorithm}[htp]
\caption{\method{} Training}
\label{alg:training}
\begin{algorithmic}[1]

\STATE \textbf{Input:} Instance distribution $\mathcal D$, group size $G$, expert weight $\alpha$, coefficient $\lambda_{\mathrm{PG}}$, clip bound $c$, and learning rates $\eta_\phi,\eta$

\STATE \textbf{Output:} Trained generator parameters $\theta$ and discriminator parameters $\phi$

\FOR{iteration $=1,2,\ldots$}
    \STATE Sample instance $s\sim\mathcal D$; encode $\{\mathbf h_i\}$; 
    $u_{ij}\leftarrow\mathrm{MLP}_{\theta_{dec}}([\mathbf h_i \,\|\, \mathbf h_j \,\|\, \mathbf e_{ij}])$

    \STATE $\tau_{neg}\sim\pi_\theta(\cdot\mid s)$;\quad 
    $\phi\leftarrow\phi-\eta_\phi\nabla_\phi R(\tau_{neg})^{2}$

    \STATE $\tau_{pos}\leftarrow$ BCC--HGS refinement of $\tau_{neg}$

    \STATE $\phi\leftarrow\phi-\eta_\phi\nabla_\phi(1-R(\tau_{pos}))^{2}$

    \STATE $\{\tau_g\}_{g=1}^{G}\sim\pi_\theta(\cdot\mid s)$ 

    \STATE $C_g\leftarrow C(\tau_g)$;\;
    $A_g\leftarrow$ Eq.~\ref{eq:advantage}

    \STATE $J_G\leftarrow\mathcal L_{\mathrm{TB}}+
    \lambda_{\mathrm{PG}}\mathcal L_{\mathrm{PG}}$

    \STATE $(\theta,Z_\theta)\leftarrow
    (\theta,Z_\theta)-\eta\nabla J_G$

\ENDFOR

\end{algorithmic}
\end{algorithm}

\begin{algorithm}[htp]
\caption{Clustering-free \method{} inference}
\label{alg:inference}
\begin{algorithmic}[1]

\STATE \textbf{Input:} {CVRP instance $s$, trained policy $\pi_\theta$, rollouts $G_s$}

\STATE \textbf{Output:} Best solution $\tau_{g^\star}$

\STATE encode $\{\mathbf h_i\}$; 
$u_{ij}\leftarrow\mathrm{MLP}_{\theta_{dec}}([\mathbf h_i \, \| \, \mathbf h_j \, \| \, \mathbf e_{ij}])$

\FOR{$g=1,\ldots,G_s$ \textbf{in parallel}}
  \FOR{$t=1,\ldots,T$}
    \STATE $a_t\sim p(a_t=v_j\mid s_t)$ 
  \ENDFOR
  \STATE $C_g\leftarrow C(\tau_g)$
\ENDFOR

\RETURN $\tau_{g^{\star}}$,\; $g^{\star}=\arg\min_{g}C_g$

\end{algorithmic}
\end{algorithm}
\textbf{Training and Inference Protocol}.
Training alternates between discriminator updates (Eq.~\ref{eq:disc}) and generator updates (Eq.~\ref{eq:total}). The BCC--HGS expert is invoked exclusively during training to provide positive samples for the discriminator. At inference, OD-Gear reverts to a clustering-free deployment, as detailed in Algorithm~\ref{alg:inference}, the  decomposition, HGS, and discriminator are omitted, and returns the lowest-cost trajectory among $G_s$ parallel rollouts.

\section{Experiments}

\textbf{Datasets.} 
We evaluate OD-Gear using standard benchmarks~\cite{kool2019attention,kim2025ant}. \emph{Synthetic Instances}: We train and test on synthetic CVRP instances with  $|V|\in\{200,500,1000\}$ and evaluate extrapolation on ultra-large instances $|V|\in\{2000,5000,\num{10000}\}$. Customer and depot coordinates are sampled from
$[0,1]^2$ with demands $d_i \sim \mathrm{Unif}\{1, \ldots, 9\}$ and capacity  $Q=50$. Each scale comprises 128 test instances. \emph{Real-world Benchmarks}: We assess generalization  on 98 instances from the CVRPLib (A, B, E, P, and Tai families), comparing against best-known solutions. Additional results are reported in Appendix~\ref{appendix_experiment_tables}: per-instance CVRPLib results appear in Tables~\ref{tab:cvrplib-A-merged}--\ref{tab:cvrplib-P-merged}, the XL benchmark~\cite{queiroga2026xl} in Tables~\ref{tab:cvrp-xl-merged-1}--\ref{tab:cvrp-xl-merged-3}, and TSP evaluations in Table~\ref{tab:merged-tsp-synthetic}. 

\textbf{Baselines.} 
We benchmark OD-Gear against a diverse set of classical and neural solvers. \emph{Classical Heuristics}:  LKH-3~\cite{helsgaun2017extension} with 100, 1000, and \num{10000} trials, and ACO~\cite{mazzeo2004ant}. \emph{Neural Solvers}: Constructive policies POMO~\cite{kwon2020pomo} and NeuOpt~\cite{ma2023learning}; the partition-based solver GLOP~\cite{ye2024glop}; and the flow-based samplers AGFN~\cite{zhang2025adversarial}, HBG~\cite{NEURIPS2025_1898054a}, and GFACS~\cite{kim2025ant}. To ensure fair comparisons, we utilize the official AGFN-based instantiation for HBG. For GFACS, we disable the local-search stage to isolate its generative performance, aligning with OD-Gear's no-local-search regime.  A suffix $s$ denotes the training scale of each learned model.


\textbf{Protocol and Metrics.} 
We follow the training and inference procedures detailed in
Algorithm~\ref{alg:training} and ~\ref{alg:inference}. OD-Gear employs sparse encoding size-adaptive partitioning ($K=|V|/5$)
and capacity-masked sampling of $G_s=100$ trajectories. 
Reported results of OD-Gear are averaged over ten independent runs from a single frozen checkpoint. Performance is measured  by the mean objective value (Obj.), the average per-instance inference latency (Time), and the optimality gap (\%). Gaps are computed relative to LKH-3(100) for synthetic instances and best-known solutions for CVRPLib and negative values indicate solutions superior to the reference. 

\textbf{Computational Environment.} All experiments are conducted on a server equipped with an NVIDIA RTX 3090\,Ti GPU and an AMD Ryzen Threadripper 3970X 32-core processor, ensuring identical hardware and software conditions across all baselines. Key hyperparameters are reported in Appendix~\ref{appendix_hyperparameter_settings}, Table~\ref{tab:hyperparameter_settings}; additional experimental details and result tables are reported in Appendix~\ref{appendix_experiment_tables}: per-instance CVRPLib results appear in Tables~\ref{tab:cvrplib-A-merged}--\ref{tab:cvrplib-P-merged}, the XL benchmark~\cite{queiroga2026xl} in Tables~\ref{tab:cvrp-xl-merged-1}--\ref{tab:cvrp-xl-merged-3}, and TSP evaluations in Table~\ref{tab:merged-tsp-synthetic}.

\begin{table*}[t]
\centering
\setlength{\tabcolsep}{4pt}
\begin{tabular}{l|rrr|rrr|rrr}
\toprule
Method & \multicolumn{3}{c|}{$|V|=200$} & \multicolumn{3}{c|}{$|V|=500$} & \multicolumn{3}{c}{$|V|=1000$} \\
 & Obj. & Gap (\%) & Time (s) & Obj. & Gap (\%) & Time (s) & Obj. & Gap (\%) & Time (s) \\
\midrule
LKH-3(100) & 28.82 & -- & 0.47 & 66.81 & -- & 1.06 & 131.77 & -- & 2.14 \\
LKH-3(1000) & 28.21 & $-2.09$ & 4.30 & 64.15 & $-3.99$ & 8.46 & 123.17 & $-6.53$ & 15.19 \\
LKH-3(\num{10000}) & 28.02 & $-2.75$ & 44.58 & 63.32 & $-5.23$ & 84.31 & 120.26 & $-8.74$ & 141.04 \\
\midrule
POMO & \textbf{29.14} & \textbf{1.13} & 2.60 & 79.41 & 18.85 & 24.09 & 191.26 & 45.14 & 192.37 \\
NeuOpt & \underline{29.46} & \underline{2.23} & 1.24 & 274.32 & 310.58 & 3.74 & -- & -- & -- \\
GLOP & 33.78 & 17.22 & \underline{0.20} & \underline{77.64} & \underline{16.21} & \textbf{0.27} & \underline{157.54} & \underline{19.55} & \textbf{0.39} \\
ACO & 97.20 & 237.29 & 1.32 & 248.33 & 271.68 & 7.25 & 503.11 & 281.80 & 26.97 \\
OD-Gear-200 & 30.10 & 4.45 & \textbf{0.13} & \textbf{68.33} & \textbf{2.27} & \underline{0.33} & \textbf{129.57} & $\textbf{-1.67}$ & \underline{0.68} \\
\midrule
GFACS & 35.17 & 22.06 & 0.16 & 78.28 & 17.16 & \underline{0.36} & 148.06 & 12.36 & 0.81 \\
AGFN-200 & \underline{30.68} & \underline{6.47} & \underline{0.14} & 69.60 & 4.17 & \textbf{0.33} & 132.57 & 0.60 & \textbf{0.68} \\
AGFN-500 & 32.24 & 11.88 & \underline{0.14} & 69.42 & 3.90 & \textbf{0.33} & 128.99 & $-2.11$ & \textbf{0.68} \\
AGFN-1000 & 32.68 & 13.40 & \underline{0.14} & 70.01 & 4.79 & \textbf{0.33} & 129.63 & $-1.63$ & \underline{0.69} \\
HBG & 30.82 & 6.94 & 0.15 & 69.90 & 4.62 & \underline{0.36} & 131.82 & 0.04 & 0.84 \\
OD-Gear-200 & \textbf{30.10} & \textbf{4.45} & \textbf{0.13} & \underline{68.33} & \underline{2.27} & \textbf{0.33} & 129.57 & $-1.67$ & \textbf{0.68} \\
OD-Gear-500 & 31.96 & 10.90 & \underline{0.14} & \textbf{67.92} & \textbf{1.65} & \textbf{0.33} & \textbf{126.79} & $\mathbf{-3.78}$ & \underline{0.69} \\
OD-Gear-1000 & 35.22 & 22.21 & \underline{0.14} & 69.15 & 3.50 & \textbf{0.33} & \underline{127.46} & $\underline{-3.27}$ & \textbf{0.68} \\
\bottomrule
\end{tabular}
\caption{Synthetic CVRP performance at three scales. 
Among non-reference methods, bold and underlined values denote the best and second-best results, respectively.  ``–” denotes failure to return a result. }
\label{tab:cvrp-synthetic}
\end{table*}

\subsection{Synthetic CVRP}
Table~\ref{tab:cvrp-synthetic} demonstrates OD-Gear's superior generalization and scalability, where negative optimality gaps indicate solutions outperforming the LKH-3(100) reference. Fixed-scale baselines suffer severe performance degradation when extrapolating: the optimality gap of POMO rises from 1.13\% ($|V|=200$) to $45.14\%$ ($|V|=1000$), NeuOpt deteriorates to $310.58\%$ at $|V|=500$ and fails at $|V|=1000$, and ACO remains above $237\%$ at all scales. Conversely, OD-Gear-200 exhibits robust extrapolation, with its gap decreasing monotonically to $-1.67\%$ at $|V|=1000$. At this scale, it produces solutions $1.7\%$ cheaper than LKH-3(100) while achieving a $3\times$ speedup over the reference (0.68s vs. 2.14s). 

OD-Gear demonstrates a superior quality--efficiency trade-off compared to multi-scale solvers. All variants maintain sub-second inference, consistently achieving the lowest optimality gaps for $|V|\in 
\{200,500,1000\}$. Notably, at $|V| = 200$, OD-Gear-200 achieves a $4.45\%$ gap, significantly outperforming AGFN-200 ($6.47\%$) and HBG ($6.94\%$). At $|V|=1000$, all OD-Gear variants yield negative gaps, outperforming  the LKH-3(100) baseline. Under the same no-local-search regime, OD-Gear-200 improves upon GFACS by 14.0–17.6 points, confirming that our gains derive from the learned constructive policy rather than post-hoc refinement. While LKH-3 variants with 1,000+ trials eventually produce cheaper solutions at $|V|\geq 500$, they incurs $22\times$--$256\times$ higher computational costs, highlighting OD-Gear's advantage in large-scale deployments.   


\subsection{Generalization on CVRPLib}

\begin{inlinetable}
\setlength{\tabcolsep}{4pt}
\begin{tabular}{lrrr}
\toprule
Method & Total Obj. & Gap (\%) & Time (s) \\
\midrule
Best known & 120120.03 & -- & -- \\
\midrule
AGFN-200 & 201463.20 & 67.72 & 43.43 \\
HBG & \underline{171100.72} & \underline{42.44} & \underline{27.96} \\
GLOP & 234860.10 & 95.52 & 70.85 \\
GFACS & 231970.83 & 93.12 & \textbf{16.79} \\
OD-Gear-200 & \textbf{143520.14} & \textbf{19.48} & 44.58 \\
\bottomrule
\end{tabular}
\par\nopagebreak
\captionof{table}{CVRPLib generalization performance.
}
\label{tab:cvrplib}
\end{inlinetable}

Table~\ref{tab:cvrplib} illustrates generalization on the real-world CVRPLib benchmark, where the OD-Gear-200 model trained solely on synthetic 200-node instances is applied without fine-tuning. OD-Gear-200 achieves the best overall balance: its total gap of $19.48\%$ represents a substantial improvement over HBG ($42.44\%$), AGFN-200 ($67.72\%$), and both GLOP and GFACS ($>93\%$). Furthermore, it attains the lowest mean gap in four of the five instance families, with HBG leading only on A-family (13.87$\%$ vs.\ 15.80$\%$). These results demonstrate that our policy effectively transfers to real-world spatial and demand distributions, although no sub-second method approaches the best-known solutions.

\subsection{Ultra-Large-Scale Extrapolation}

\begin{inlinetable}
\setlength{\tabcolsep}{1.9pt}
\begin{tabular}{l|rr|rr|rr}
\toprule
 & \multicolumn{2}{c|}{$|V|=2000$} & \multicolumn{2}{c|}{$|V|=5000$} & \multicolumn{2}{c}{$|V|=\num{10000}$} \\
Method & Obj. & Time & Obj. & Time & Obj. & Time \\
\midrule
LKH-3(1000) & 243.57 & 28.49 & 631.60 & 47.00 & 1292.65 & 65.85 \\
POMO-100 & 648.76 & 3.35 & 2146.48 & 76.47 & -- & -- \\
GLOP & 287.39 & \textbf{1.01} & 712.80 & \textbf{1.76} & 1319.47 & \textbf{3.94} \\
AGFN-200 & 254.05 & 7.55 & \underline{587.56} & 19.53 & \textbf{1145.80} & 38.36 \\
GFACS & 284.33 & 1.50 & 682.98 & \underline{4.78} & 1349.50 & 10.96 \\
HBG & \underline{250.62} & 1.71 & 603.99 & 5.42 & \underline{1166.56} & 11.58 \\
\midrule
OD-Gear-200 & \textbf{249.93} & \underline{1.46} & \textbf{582.39} & 4.88 & 1210.10 & \underline{9.74} \\
\bottomrule
\end{tabular}
\par\nopagebreak
\captionof{table}{Ultra-large synthetic CVRP. Suffixes denote training scales and Time is seconds per instance. }
\label{tab:cvrp-ultra}
\end{inlinetable}

Table~\ref{tab:cvrp-ultra} evaluates performance on ultra-large instances, requiring $10\times$--$50\times$ extrapolation from small-scale checkpoints. OD-Gear-200 surpasses LKH-3(1000) at $|V|=5000$ (582.39 vs.\ 631.60) and $|V|=\num{10000}$ (1210.10 vs.\ 1292.65), while delivering $9.6\times$ and $6.8\times$ speedups, respectively. At $|V|=2000$, it remains within $2.7\%$ of the reference with a $19\times$ speedup and attains the lowest objective in the table at $|V|=5000$. Compared to baselines under the same no-local-search regime,  GFACS incurs 11--17$\%$ higher costs at comparable runtimes, whereas GLOP is faster, it yields 9--22$\%$ worse solutions. POMO-100 fails to scale to $|V|=\num{10000}$. Although AGFN-200 and HBG achieve lower objectives at $|V|=\num{10000}$, they require $3.9\times$ and $1.2\times$ the runtime of OD-Gear. This superior accuracy-latency trade-off at the \num{10000}-node scale validates the robustness of our expert-guided training scheme in extreme extrapolation scenarios.


\subsection{Ablation Study}

We evaluate OD-Gear against three controlled variants: \emph{w/o GroupSampling}, which removes the group-relative advantage and the auxiliary group-sampling objective; \emph{w/o Decomposition}, which disables BCC in the training expert; \emph{w/ Transformer}, which replaces the GAT backbone with a standard Transformer architecture. By isolating these factors, we ensure a rigorous analysis of our architectural and training innovations, with full results summarized in Table~\ref{tab:ablation}.


\textbf{Component Ablation}.
\emph{Impact of Group Sampling.} At the matched training--evaluation scales, group sampling improves on the \emph{w/o GroupSampling} baseline of Table~\ref{tab:ablation} at $|V|=200$ (4.45$\%$ vs.\ 5.62$\%$ ) and $|V|=1000$ (-3.27$\%$ vs.\ -3.16$\%$), both with $p<10^{-12}$ (ten-run Welch's test). At $|V|=500$, performance is statistically indistinguishable (1.65$\%$ vs.\ 1.66$\%$, $p=0.57$). Moreover, it also wins the 500-trained model's 1000-node cell (-3.78$\%$ vs.\ -3.70$\%$). In cross-scale settings, the \emph{w/o GroupSampling} variant retains a 0.4--4.6 point advantage in five configurations. This indicates that while group sampling sharpens accuracy at the training scale, it may slightly trade off cross-scale transferability.

\begin{inlinetable}
\setlength{\tabcolsep}{6pt}
\begin{tabular}{lrrr}
\toprule
 & \multicolumn{3}{c}{Gap (\%)} \\
Method & $|V|{=}200$ & $|V|{=}500$ & $|V|{=}1000$ \\
\midrule
OD-Gear-200 & \textbf{4.45} & \underline{2.27} & $\underline{-1.67}$ \\
\;\;w/o GroupSampling & \underline{5.62} & \textbf{1.70} & $\mathbf{-2.77}$ \\
\;\;w/o Decomposition & 11.79 & 4.84 & $-1.30$ \\
\;\;w/ Transformer & 15.85 & 8.59 & 1.66 \\
\midrule
OD-Gear-500 & \underline{10.90} & \textbf{1.65} & $\mathbf{-3.78}$ \\
\;\;w/o GroupSampling & \textbf{10.20} & \underline{1.66} & $\underline{-3.70}$ \\
\;\;w/o Decomposition & 12.25 & 4.59 & $-1.78$ \\
\;\;w/ Transformer & 17.34 & 9.63 & 2.67 \\
\midrule
OD-Gear-1000 & 22.21 & \underline{3.50} & $\mathbf{-3.27}$ \\
\;\;w/o GroupSampling & 17.63 & \textbf{3.14} & $\underline{-3.16}$ \\
\;\;w/o Decomposition & \textbf{12.85} & 4.81 & $-1.69$ \\
\;\;w/ Transformer & \underline{16.21} & 10.93 & 4.45 \\
\bottomrule
\end{tabular}
\par\nopagebreak
\captionof{table}{Ablation study on synthetic CVRP.
}
\label{tab:ablation}
\end{inlinetable}

\emph{Impact of Decomposition.} Excluding expert clustering degrades performance in eight of nine cells by 0.4--7.3 points, with the most significant drop occurring 
at the training scale of OD-Gear-200 (4.45$\%$ to 11.79$\%$). The exception is the 1000-trained model at $|V|=200$, where the full model itself transfers poorly (22.21$\%$ vs.\ 12.85$\%$).

\emph{Architectural Comparison.} Replacing the GAT backbone with a Transformer reduces performance in eight of nine cells settings. Notably, this is the only variant that exhibits a positive gap at $|V|=1000$ (1.66$\%$--4.45$\%$), confirming the advantage of sparse edge-aware attention in modeling the coupled spatial and capacity structure of the CVRP.

\emph{Discussion.} Our results confirm the sparse structural encoder and decomposition expert as primary performance drivers, while group sampling further improves training-scale accuracy at no additional inference cost.

\textbf{Sensitivity Analysis}.
\begin{inlinetable}
\setlength{\tabcolsep}{4.5pt}
\begin{tabular}{rrrrr}
\toprule
 & \multicolumn{3}{c}{Gap (\%)} & Time (s) \\
$G_s$ & $|V|{=}200$ & $|V|{=}500$ & $|V|{=}1000$ & $|V|{=}1000$ \\
\midrule
10 & 5.08 & 2.87 & $-1.32$ & 0.66 \\
50 & 4.57 & 2.44 & $-1.57$ & 0.67 \\
100 & 4.45 & 2.27 & $-1.67$ & 0.68 \\
300 & 4.21 & 2.09 & $-1.79$ & 0.69 \\
500 & 4.11 & 1.99 & $-1.85$ & 0.70 \\
700 & \underline{4.06} & \underline{1.94} & $\underline{-1.89}$ & 0.75 \\
1000 & \textbf{3.98} & \textbf{1.88} & $\mathbf{-1.92}$ & 0.84 \\
\midrule
$K$ & & & & \\
\midrule
20 & 8.25 & 7.57 & 12.78 & 0.68 \\
30 & 5.22 & 1.51 & \underline{1.50} & 0.66 \\
40 & \textbf{4.88} & \textbf{0.48} & \textbf{0.38} & 0.67 \\
50 & \underline{5.11} & \underline{0.60} & 2.51 & 0.67 \\
60 & 5.39 & 0.71 & 1.96 & 0.66 \\
\bottomrule
\end{tabular}
\par\nopagebreak
\captionof{table}{OD-Gear-200 sensitivity. 
}
\label{tab:sensitivity}
\end{inlinetable}
\emph{Effect of Inference Group Size.} Increasing $G_s$ yields diminishing returns on solution quality (Table~\ref{tab:sensitivity}, top).
Expanding $G_s$ from 10 to the default 100 gains 0.35--0.63 points, whereas a further $10\times$ increase to 1000 adds only 0.25--0.47 points while increasing the 1000-node latency from 0.68\,s to 0.84\,s. The default thus operates near the Pareto optimal point of the quality–latency curve.

\emph{Effect of the Sparse Neighborhood.} Setting $K=20$ over-sparsifies the graph, degrading the gap by 3.4--12.4 points compared to $K=40$, whereas quality remains stable for $K\in[30,60]$ (Table~\ref{tab:sensitivity}, bottom). Since  this sweep fixes $K$ across scales, these results are not directly comparable to the size-adaptive main configuration beyond 200 nodes.

\section{Conclusion}
We proposed OD-Gear, an adversarial learning framework that effectively addresses scalability bottlenecks in large-scale CVRP. By distilling ``divide-and-conquer” logic from a hybrid HGS-BCC expert into a GAT-based generative policy and leveraging group-relative advantages to promote solution diversity, OD-Gear enables high-quality, clustering-free inference on massive graphs. Empirical results show that OD-Gear achieves  SOTA real-time performance across most benchmarks and maintains high efficiency at the 10,000-node scale. This paradigm successfully  bridges the gap between classical heuristic precision and scalable neural optimization. Future work will extend this framework to multi-constraint variants, such as CVRP with time windows, and further refine its inductive biases for diverse graph topologies.

\makeatletter
\par
\if@firstcolumn\newpage\null\newpage\else\null\newpage\fi
\makeatother
\appendix

\setcounter{figure}{1}
\setcounter{table}{0}
\section{Formal Mathematical Formulations}
\label{appendix_math_formulation}

\subsection{Mathematical Formulation for the CVRP}\label{appendix_cvrp_ilp}
The capacitated vehicle routing problem (CVRP) is a fundamental challenge in combinatorial optimization. This section provides a formal integer linear programming (ILP) formulation. The objective is to design optimal delivery routes that minimize total travel cost while satisfying vehicle capacity constraints. Following the classical flow-based formulation, we define the problem on a graph
$G(V,E)$, where $V=\{v_0\} \cup V_c$ represents the depot and the set of customers, respectively. 
We introduce binary decision variables $x_{ij} \in \{0,1\}$, where $x_{ij}=1$ if a vehicle traverses the edge from node $i$ to node $j$, and 0 otherwise. Meanwhile, a continuous variable $u_i$ represents the cumulative load upon arrival at node $i$, serving to both enforce capacity limits and eliminate subtours. The CVRP is formulated as follows:
{
\begin{alignat}{2}
\min \quad
& \sum_{i \in V}
  \sum_{\substack{j \in V \\ j \neq i}}
  c_{ij}x_{ij}
&&
\tag{1a}\label{1a}
\\
\text{s.t.}\quad
& \sum_{\substack{j \in V \\ j \neq i}}
  x_{ij}=1,
&& \forall i\in V_c,
\tag{1b}\label{1b}
\\
&
\sum_{\substack{i \in V \\ i \neq j}}x_{ij}
-\sum_{\substack{k \in V \\ k \neq j}}x_{jk}=0,
&& \forall j\in V_c,
\tag{1c}\label{1c}
\\
&
\sum_{j\in V_c}x_{0j}\le K,
&&
\tag{1d}\label{1d}
\\
&
u_i-u_j+Qx_{ij}\le Q-d_j,
&& \forall i,j\in V_c,\ i\neq j,
\tag{1e}\label{1e}
\\
&
d_i\le u_i\le Q,
&& \forall i\in V_c,
\tag{1f}\label{1f}
\\
&
x_{ij}\in\{0,1\},
&& \forall i,j\in V,\ i\neq j,
\tag{1g}\label{1g}
\\
&
u_i\ge 0,
&& \forall i\in V_c.
\tag{1h}\label{1h}
\end{alignat}
}

The objective function \eqref{1a} minimizes the total travel cost (e.g., total distance or time) for the entire fleet. constraints \eqref{1b} and  \eqref{1c} enforce service uniqueness and flow conservation, ensuring that each customer is visited exactly once and that vehicles maintain path continuity. Constraints \eqref{1d} restrict the total number of vehicles departing from the depot not to exceed the fleet size $K$. To prevent the formation of isolated subtours and enforce the vehicle capacity $Q$, we utilize the Miller-Tucker-Zemlin (MTZ) formulation in \eqref{1e}, where the auxiliary variables $u_i$ track the cumulative demand along each route. Finally, constraints \eqref{1f}-\eqref{1h} define the binary requirements for the routing variables and the feasible ranges for the continuous load variables, ensuring a rigorous mathematical representation of the problem.

\subsection{Mathematical Formulation for the TSP}\label{appendix_tsp_formulation}
The traveling salesman problem (TSP) is a foundational task in combinatorial optimization and can be viewed as a specialized case of the CVRP. In this section, we provide a formal integer linear programming (ILP) formulation of the TSP to establish the theoretical basis for our subsequent experimental  analysis.

The objective of the TSP is to identify the minimum-cost Hamiltonian cycle that visits each node in a set $V = \{1, \dots, n\}$ exactly once before returning to the origin. 
Formally, let $G=(V, E)$ be a complete graph where $c_{ij}$ represents the cost associated with edge $(i, j) \in E$. We define the binary decision variable $x_{ij} \in \{0,1\}$, which equals 1 if the path traverses directly from node $i$ to node $j$, and 0 otherwise. 
To prevent the formation of isolated subtours, we employ the Miller-Tucker-Zemlin (MTZ) formulation, introducing auxiliary continuous variables $u_i$ to track the sequence of visitation. The TSP is formulated as follows:
\allowdisplaybreaks
\begin{alignat}{2}
    \min \quad & \sum_{i \in V} \sum_{\substack{j \in V \\ j \neq i}} c_{ij} x_{ij} \tag{2a} \label{2a} \\
    \text{s.t.} \quad & \sum_{\substack{j \in V \\ j \neq i}} x_{ij} = 1, && \forall i \in V, \tag{2b} \label{2b} \\
    & \sum_{\substack{i \in V \\ i \neq j}} x_{ij} = 1, && \forall j \in V, \tag{2c} \label{2c} \\
    & u_i - u_j + n x_{ij} \le n - 1, && \forall i,j \in \{2, \dots, n\}, i \ne j, \tag{2d} \label{2d} \\
    & 1 \le u_i \le n - 1, && \forall i \in \{2, \dots, n\}, \tag{2e} \label{2e} \\
    & x_{ij} \in \{0,1\}, && \forall i,j \in V, i \ne j. \tag{2f} \label{2f}
\end{alignat}%
The objective function 
\eqref{2a} minimizes the total travel cost of the Hamiltonian cycle. Constraints \eqref{2b} and \eqref{2c} are the degree constraints, which respectively ensure that exactly one edge departs from and enters every node; together, they guarantee that each city is visited exactly once. Constraints \eqref{2d} denote the Miller-Tucker-Zemlin (MTZ) subtour elimination criteria. It enforces that if a path exists from $i$ to $j$ (i.e., $x_{ij}=1$), then the visitation order must satisfy $u_j \ge u_i + 1$.  these constraints preclude the formation of closed loops (subtours) that do not include the designated starting node (node 1). Finally, constraints \eqref{2e} and  \eqref{2f} 
define the feasible domains for the auxiliary sequence variables
$u_i$ and the binary decision variables $x_{ij}$.

\section{Graph Attention Network (GAT) Encoder}\label{appendix_gat_details}
In the OD-Gear framework, the generator adopts a graph attention networks (GAT) as its core policy architecture. GATs are particularly well-suited to the CVRP because they overcome the limitations of standard graph convolutional networks (GCNs). In particular, GATs dynamically weight interactions among customer nodes through an attention mechanism, enabling the model to capture heterogeneous node features and adapt to varying graph topologies. These properties are critical for routing problems, where long-range spatial dependencies and demand heterogeneity play a central role. This section introduces the construction of input features, presents the mathematical formulation of the attention mechanism, and describes the multi-head attention aggregation strategy.

For a given CVRP instance, the input to the GAT is a node feature matrix $X=\{\mathbf{x}_1, \mathbf{x}_2, ..., \mathbf{x}_N\}$, where each $\mathbf{x}_i \in \mathbb{R}^F$ represents the raw
feature vector of node $i$. In our implementation, $\mathbf{x}_i$ contains the normalized 2D coordinates $(x_i, y_i)$ and the customer demand $d_i$. To project these features into a high-dimensional representation space, we apply a learnable linear transformation via a shared weight matrix $W_{node} \in \mathbb{R}^{D_{units} \times F}$. The initial hidden state $\mathbf{h}_i^{(0)} = \sigma(W_{node}\mathbf{x}_i + b_{node})$. This projection maps raw inputs into a unified latent space. It enables the model to capture complex spatial and demand-based relationships between nodes in subsequent attention layers.

The primary innovation of GAT 
is its self-attention mechanism, 
which enables nodes to dynamically assign aggregation weights based on neighborhood features, rather than relying on a fixed graph Laplacian matrix. Following Veličković et al. \cite{velivckovic2018graph}, for a node $i$ and its neighbor $j \in \mathcal{N}_i$, the attention coefficient $s_{ij}$ quantifies the relative importance of node $j$ to node $i$. This paper adopts an additive attention mechanism:
\begin{equation}
    s_{ij} = \text{LeakyReLU}(\vec{a}^T [W_{node}\mathbf{x}_i || W_{node}\mathbf{x}_j]),
    \tag{3}
\end{equation}
where $||$ denotes the vector concatenation operation and $\vec{a} \in \mathbb{R}^{2D_{units}}$ is a learnable weight vector of a single-layer feedforward neural network. To normalize these coefficients into a probability distribution, we apply the softmax function: $\alpha_{ij} = \text{softmax}_j(s_{ij})$. This weight $\alpha_{ij}$ explicitly dictates the contribution of node $j$'s feature during the information passing process to node $i$.

To stabilize training and capture diverse feature subspaces, such as  geographic proximity or demand-capacity correlations, we utilize  a multi-head attention mechanism. With $K$ independent attention heads,  the intermediate layers employ a concatenation strategy: $\mathbf{h}_i' = ||_{k=1}^K \sigma(\sum_{j \in \mathcal{N}_i} \alpha_{ij}^k W^k \mathbf{x}_j)$. For the output layer, an averaging strategy is used to produce a fixed-dimension embedding. Compared to fully connected Transformer-based architectures with $O(N^2)$ complexity, GAT offers significant advantages for large-scale CVRP. By restricting attention to a $k$-nearest neighbor ($k$-NN) graph, the complexity is reduced to $O(kN)$, drastically improving inference speed. Furthermore, the explicit modeling of graph topology provides GAT with strong permutation invariance, a fundamental property for routing problems where solution quality must be independent of node indexing.

\section{Motivation Behind the Group-Relative Advantage}
\label{appendix_grpo}
The design of the group-relative advantage is motivated by the need to mitigate the high variance of single-trajectory estimates and foster solution diversity. Specifically, for each training instance, OD-Gear samples a diverse set of candidate trajectories and computes a group-relative advantage based on the cost deviation of each trajectory from the group mean. Incorporating this advantage into the trajectory-balance (TB) objective enhances the model's ability to discriminate  between solutions beyond the expert-guided discriminator reward, effectively 
allocating more training flow to lower-cost trajectories while penalizing inferior ones. 
Concurrently, an auxiliary policy-gradient (PG) objective directly increases the generation probabilities of trajectories that outperform the group average and decreases those of inferior trajectories.
Consequently, TB objective shapes the overall 
distribution of high-quality candidate solutions, whereas the PG term provides targeted reinforcement to superior trajectories through direct policy updates. By leveraging this within-group relative information, OD-Gear achieves superior solution quality while maintaining effective exploration across the diverse solution space.

\section{Standardized Notations, Hyperparameters, and Baselines}
\label{appendix_notations_baselines}

\subsection{Summary of Mathematical Notations}
\label{appendix_notation_table}

To facilitate clarity in our mathematical formulation, Table~\ref{tab:notations} summarizes the primary notations and their corresponding definitions used throughout this paper.

\subsection{Hyperparameter Settings}
\label{appendix_hyperparameter_settings}
Table~\ref{tab:hyperparameter_settings} summarizes the hyperparameters for the OD-Gear model, group-sampling strategy, and sensitivity analyses to ensure reproducibility.

\newcolumntype{C}[1]{>{\centering\arraybackslash}p{#1}}
\begin{table*}[t]
    \centering
    \renewcommand{\arraystretch}{1.1}

    \small 

    \begin{tabular}{l C{0.6\linewidth}}
        \toprule
        \textbf{Symbol} & \textbf{Definition and Description} \\
        \midrule
        \multicolumn{2}{c}{\textit{Problem Definition}} \\
        $G(V, E)$ & Complete graph for CVRP or TSP, where $V$ is the node set and $E$ is the edge set \\
        $V_c$ & Set of customer nodes $\{v_1, \dots, v_N\}$ \\
        $v_0$ & Depot node \\
        $N$ & Total number of customer nodes \\
        $c_{ij}$ & Euclidean distance or travel cost associated with edge $(i, j)$ \\
        $d_i$ & Demand of node $i$ \\
        $K$ & Total number of available vehicles \\
        $Q$ & Maximum capacity of a vehicle \\
        $x_{ij}$ & Binary decision variable; 1 if a vehicle travels directly from $i$ to $j$, 0 otherwise \\
        $u_i$ & Auxiliary variable representing accumulated load or visitation order at node $i$ \\
        \midrule
        \multicolumn{2}{c}{\textit{OD-Gear Model}} \\
        $\pi_{\theta}$ & Generator policy network parameterized by $\theta$ \\
        $D_{\phi}$ & Discriminator network parameterized by $\phi$ \\
        $P_E$ & Expert policy distribution (constructed via HGS and BCC decomposition) \\
        $S_{\text{student}}$ & Student solution generated by the generator \\
        $S_{\text{expert}}$ & Expert solution generated by the expert system \\
        $h_i$ & High-dimensional embedding vector of node $i$ (generated by GAT encoder) \\
        $\alpha_{ij}$ & Attention coefficient representing the importance weight of node $j$ to node $i$ \\
        $R(S)$ & Surrogate reward signal for solution $S$ defined by the discriminator output \\
        $\mathcal{L}_{D}$ & Discriminator Loss Function \\
        $\{\tau_g\}_{g=1}^{G_s}$ & Group of candidate trajectories sampled from the generator policy $\pi_{\theta}$ \\
        $\tau_g$ & The $g$-th sampled trajectory generated by the policy network \\
        $C_g$ & Cost value of the $g$-th sampled trajectory $\tau_g$ \\
        $A_g$ & Group-relative advantage of trajectory $\tau_g$\\
        $\bar{C}$ & Mean cost of sampled trajectories within the group \\
        $\sigma_C$ & Standard deviation of trajectory costs within the sampled group \\
        $\mathcal{L}_{PG}$ & Policy gradient loss based on group-relative advantages \\
        $\mathcal{L}_{TB}$ & Trajectory Balance Loss \\
        \bottomrule
    \end{tabular}
    \caption{Summary of primary  mathematical notations.}
    \label{tab:notations}
\end{table*}

\begin{table*}[t]
    \centering
    \small
    \setlength{\tabcolsep}{3.5pt}
    \renewcommand{\arraystretch}{1.08}

    \begin{tabularx}{\linewidth}{
        @{}
        p{0.22\linewidth}
        C{0.04\linewidth}
        C{0.18\linewidth}
        >{\raggedright\arraybackslash}X
        @{}
    }
        \toprule
        \textbf{Hyperparameter}
        & \textbf{Symbol}
        & \textbf{Value}
        & \textbf{Description} \\
        \midrule

        \multicolumn{4}{c}{\textit{Original Model and Inference Settings}} \\
        \midrule

        Number of GAT layers
        & $L$
        & $4$
        & Number of stacked GAT layers in the structural encoder. \\

        Hidden embedding dimension
        & $D_{\mathrm{units}}$
        & $64$
        & Dimension of the node and edge embeddings. \\

        Inference exploration rate
        & $\rho$
        & $0.05$
        & Probability of stochastic action sampling during inference rollouts. \\

        Target subproblem size
        & $m$
        & $50$
        & Target size of each subproblem generated by BCC. \\

        Generator learning rate
        & $\eta$
        & $5\times10^{-4}$
        & Learning rate for updating $\theta$ and $Z_{\theta}$. \\

        Discriminator learning rate
        & $\eta_{\phi}$
        & $1\times10^{-3}$
        & Learning rate for updating discriminator parameters $\phi$. \\

        Sparse neighborhood size
        & $K$
        & $|V|/5$
        & Size-adaptive neighborhood used by the sparse encoder. \\

        Inference rollout number
        & $G_s$
        & $100$
        & Number of parallel trajectories sampled during inference. \\

        \midrule
        \multicolumn{4}{c}{\textit{Group-Sampling Settings}} \\
        \midrule

        Training group size
        & $G$
        & $20$
        & Number of trajectories sampled for each training instance. \\

        Numerical stability constant
        & $\epsilon$
        & $10^{-8}$
        & Prevents division by zero in advantage normalization. \\

        Advantage clip bound
        & $c$
        & $5$
        & Clips the normalized advantage to $[-5,5]$. \\

        Expert and advantage weight
        & $\alpha$
        & $500 \rightarrow 2000$
        & Logarithmic schedule for CVRP-200; endpoints scale with instance size. \\

        Policy-gradient loss weight
        & $\lambda_{\mathrm{PG}}$
        & $0.1$
        & Balances the auxiliary PG loss and the TB loss. \\

        Top-$k$ trajectory filter
        & $k$
        & $0$
        & Uses the full sampled group without top-$k$ filtering. \\

        \midrule
        \multicolumn{4}{c}{\textit{Sensitivity Analysis Settings}} \\
        \midrule

        Inference rollout sweep
        & $G_s$
        & $\{10,50,100,300,$
          \newline $500,700,1000\}$
        & Candidate rollout numbers; the default is $G_s=100$. \\

        Sparse neighborhood sweep
        & $K$
        & $\{20,30,40,50,60\}$
        & Fixed neighborhood sizes used in the sensitivity study. \\

        \bottomrule
    \end{tabularx}
\par\smallskip
\raggedright \noindent \textbf{Note}: Scale-specific configurations: CVRP-1000 model is trained with group size $G=10$, CVRPLib evaluation uses $G_s=200$ rollouts with a $K=|V|/4$ neighborhood, and ultra-large instances ($|V|\geq 5000$) use the size-adaptive default $K=|V|/10$. 
    \caption{OD-Gear Hyperparameter settings.
    }
    \label{tab:hyperparameter_settings}
\end{table*}
\subsection{Baseline Algorithm Descriptions}
\label{appendix_baselines}
To validate the performance of OD-Gear, we compare our method against six representative baselines, spanning traditional heuristics, constructive neural solvers, and GFlowNet-based generative methods. This section provides an overview of their core features and applicability. Unless otherwise specified, all baselines utilize the default configurations reported in their respective publications.
\paragraph{LKH-3.} 
LKH-3 \citep{helsgaun2000effective} is a classical heuristic algorithm based on an improved version of the Lin-Kernighan-Helsgaun heuristic. It is widely regarded as a strong baseline in the VRP field capable of generating near-optimal solutions. However, as a purely heuristic method lacking neural dependencies, its computational time grows exponentially with problem size, making it impractical for instances with 10,000 nodes within reasonable time limits.

\paragraph{ACO.} 
ACO \cite{mazzeo2004ant} is a metaheuristic inspired by the collective foraging behavior of ant colonies, designed to identify high-quality solutions through iterative pheromone updates and probabilistic path construction. While robust across various combinatorial tasks, ACO is characterized by high computational complexity and slow convergence rates. On large-scale instances, the algorithm frequently exhibits premature convergence to local optima and requires extensive parameter tuning, making it significantly less efficient than end-to-end neural architectures for real-time deployment.

\paragraph{POMO.} 
POMO \citep{kwon2020pomo} is a representative neural constructive solver based on the Transformer architecture. It utilizes the REINFORCE algorithm and enhances solution quality through policy augmentation via multiple optimal rollouts. While POMO achieves fast inference, its generalization ability degrades significantly on large-scale instances (beyond 100 nodes), and it is computationally limited by the quadratic complexity of its self-attention mechanism.

\paragraph{NeuOpt.} 
NeuOpt \citep{ma2023learning} is a neural improvement heuristic that enhances constructive solvers through a flexible  $k$-opt neural search mechanism designed to refine local route structures. While NeuOpt achieves competitive accuracy on small-scale benchmarks, it incurs substantial computational overhead and faces significant scalability challenges, often becoming impractical for instances exceeding 1,000 nodes.

\paragraph{GFACS.} 
GFACS \citep{kim2025ant} is a hybrid solver that combines GFlowNet with ant colony optimization (ACO), leveraging the generative capability of GFlowNet to guide the search process. While this synergy enhances solution diversity, the framework’s heavy reliance on ACO for iterative refinement introduces significant computational overhead. Consequently, GFACS suffers from high inference latency and exhibits limited scalability on large-scale instances, where the stochastic search phase becomes a bottleneck compared to end-to-end neural solvers.

\paragraph{AGFN.}  AGFN \cite{zhang2025adversarial} integrates  integrates adversarial training with flow matching to provide denser training signals and enhance solution diversity compared to traditional RL approaches. However, AGFN improves solution diversity, it lacks the structural inductive biases and expert regularization required for large-scale optimization. By neglecting local signals in favor of holistic objectives, such solvers fail to capture fine-grained structures, resulting in suboptimal performance ceilings and reduced training stability.

\textbf{GLOP.} ~GLOP \cite{ye2024glop}  is a hierarchical framework that addresses massive routing by integrating global partitioning with local construction. The architecture decomposes instances into a hierarchy of subproblems. Specifically, the model combines nonautoregressive partitioning with autoregressive solvers to address TSPs and shortest Hamiltonian path problems, balancing scalability and route quality. While GLOP delivers strong real-time performance on massive graphs, its multi-stage pipeline necessitates higher modeling complexity relative to end-to-end neural solvers.

\textbf{HBG.} ~HBG \cite{zhang2026hybrid} is a hybrid GFlowNet framework that unifies trajectory balance (TB) and detailed balance (DB) to jointly optimize
global trajectory-level and local state-transition objectives. By adaptively balancing these complementary objectives, HBG introduces a depot-guided inference strategy tailored for depot-centric problems like CVRP, while maintaining compatibility with depot-free tasks such as TSP. HBG can be integrated into existing GFlowNet-based solvers, such as AGFN and GFACS, to enhance solution quality and generalization. 
However, the introduction of an additional balance objective and adaptive integration mechanism increases training complexity compared to standard TB-based GFlowNet methods.


\section{Supplementary Experimental Details}
\label{appendix_experiment_tables}

This appendix provides comprehensive experimental details to facilitate reproducibility, including full hyperparameter configurations, additional sensitivity analyses, and the complete set of experimental results (Tables~\ref{tab:merged-cvrp-synthetic}--\ref{tab:merged-tsp-synthetic}). 

\textbf{Evaluation Protocol.} 
Following the main paper, the evaluation protocol is defined as follows:
\begin{itemize}
    \item For synthetic instances, the optimality gap is computed relative to LKH-3(100).
    \item For CVRPLib, gaps are computed relative to the best-known solutions (BKS).
    \item Negative gaps indicate solutions that are cheaper than the reference.
    \item Time represents the average per-instance inference latency in seconds.
    \item A suffix $s$ denotes the training scale of a learned solver.
\end{itemize}


\textbf{Naming Conventions.}
\begin{itemize}
    \item \emph{Ablations}: Consistent with the main paper, \textbf{OD-Gear-$s$ w/o Group Sampling} removes the group-relative advantage and the auxiliary group-sampling objective; \textbf{w/o Decomposition} disables barycenter clustering in the training expert; and \textbf{w/ Transformer} replaces the GAT backbone. Note that CVRP w/o Group Sampling rows are ten-run means, whereas their TSP counterparts retain the original single-run record. 
    \item \emph{Baselines}: \textbf{GFACS-noLS} denotes GFACS with its local-search stage disabled, the configuration reported as ``GFACS'' in the main paper, while rows labeled \textbf{GFACS} or \textbf{GFACS-200} keep local search enabled and are provided for completeness.
    \item  \emph{Notations}: $\hat{N}$ denotes the inference group size ( referred to $G_s$ in the main paper), and $K$ is the sparse neighborhood size ($k_{\mathrm{sparse}}$), whose size-adaptive default is $K=|V|/5$.
\end{itemize}


\textbf{Synthetic CVRP.} Table~\ref{tab:merged-cvrp-synthetic} extends the synthetic-CVRP comparison from the main paper with the full objective values,  additional results for AGFN-500/1000, and local-search-enabled GFACS rows. The OD-Gear-200/500/1000 results  are reported as ten-run means using frozen-checkpoint. Gaps for all newly added methods are recomputed against LKH-3(100) to maintain consistency with the paper's convention.

\textbf{CVRPLib and TSPLib aggregates.} Table~\ref{tab:merged-lib} reports the aggregate objective values, wall-clock time, and gap across  the 98 CVRPLib instances ( comprising the  A, B, E, P, and Tai families), together with the corresponding TSPLib aggregate. The OD-Gear-200 CVRPLib gap uses the 98-instance aggregate best-known total 120120.03.

\textbf{Ultra-large-scale CVRP.} Table~\ref{tab:merged-large-cvrp} presents the ultra-large-scale extrapolation results, including the local-search-enabled GFACS-200 row omitted from the main paper.

\textbf{Ablations.} Table~\ref{tab:merged-ablation-odgear} reports the controlled OD-Gear, w/o Decomposition, and w/ Transformer ablations used in the main paper. All controlled OD-Gear entries are ten-run means evaluated  without inference-time local search, with gaps recomputed relative to LKH-3(100).

\textbf{Sensitivity analyses.} Tables~\ref{tab:merged-hyper-n-oddeal} and~\ref{tab:merged-hyper-n-grpo} give the full parameter sweeps of the inference group size $\hat{N}$ ($G_s$) for OD-Gear-200 w/o Group Sampling and OD-Gear-200, respectively. Tables~\ref{tab:merged-hyper-k-oddeal} and~\ref{tab:merged-hyper-k-grpo} present the sweeps over the sparsification degree $K$. For OD-Gear $\hat{N}$ sweeps, results are derived from the frozen-checkpoint re-evaluation used by the main paper; they are complete up to $\hat{N}=1000$, partial available at $\hat{N}=1100$ (where the $|V|=1000$ column is missing), while  settings above 1100 were not re-run. For the $K$ sweeps, $K$ is fixed in the range of $20$--$60$ across all evaluation scales, whereas the main rows use the size-adaptive default $K=|V|/5$ (i.e., $40$, $100$, $200$ for $|V|=200$, $500$, $1000$, respectively). Consequently, only the $|V|=200$ column (where the adaptive setting $K=40$ coincides with the sweep point) aligns with the main tables. Furthermore, 
the $K=50$/$60$ rows at $|V|=1000$ are non-monotone because a fixed small $K$ becomes increasingly sub-optimal as the instance grows.

\textbf{Per-instance results.} Tables~\ref{tab:cvrplib-A-merged}--\ref{tab:cvrplib-P-merged} report the per-instance best, mean, gap, and time on the five CVRPLib families.  Tables~\ref{tab:cvrp-xl-merged-1}--\ref{tab:cvrp-xl-merged-3} summarize  the per-instance results on the 100-instance XL CVRP benchmark.

\textbf{Synthetic TSP.} Table~\ref{tab:merged-tsp-synthetic} reports the synthetic TSP comparison deferred from the main paper. OD-Gear results are reported as ten-run means, and gaps for the added methods are recomputed against LKH-3(100).

\begin{table*}[!tp]
\centering
\resizebox{\textwidth}{!}{%
\begin{tabular}{l|ccc|ccc|ccc}
        \toprule \toprule
        Method & \multicolumn{3}{c|}{$|V|=200$} & \multicolumn{3}{c|}{$|V|=500$} & \multicolumn{3}{c}{$|V|=1000$} \\
        ~ & Obj. & Gap (\%) & Time (s) & Obj. & Gap (\%) & Time (s) & Obj. & Gap (\%) & Time (s) \\
        \midrule
        LKH-3(100) & 28.816896 & - & 0.47 & 66.812744 & - & 1.06 & 131.771685 & - & 2.14 \\
        LKH-3(1000) & 28.213308 & -2.09 & 4.30 & 64.147637 & -3.99 & 8.46 & 123.169293 & -6.53 & 15.19 \\
        LKH-3(10,000) & 28.024617 & -2.75 & 44.58 & 63.320593 & -5.23 & 84.31 & 120.260032 & -8.74 & 141.04 \\
        POMO & 29.141700 & 1.13 & 2.60 & 79.408600 & 18.85 & 24.09 & 191.255300 & 45.14 & 192.37 \\
        NeuOpt & 29.459557 & 2.23 & 1.24 & 274.321472 & 310.58 & 3.74 & - & - & - \\
        GLOP & 33.779831 & 17.22 & 0.20 & 77.640152 & 16.21 & 0.27 & 157.538406 & 19.55 & 0.39 \\
        ACO & 97.195656 & 237.29 & 1.32 & 248.330895 & 271.68 & 7.25 & 503.110024 & 281.80 & 26.97 \\
        GFACS & 28.507015 & -1.08 & 3.25 & 64.381401 & -3.64 & 8.22 & 122.071861 & -7.36 & 15.82 \\
        AGFN-200 & 30.680285 & 6.47 & 0.14 & 69.596748 & 4.17 & 0.33 & 132.565826 & 0.60 & 0.68 \\
        AGFN-500 & 32.240875 & 11.88 & 0.14 & 69.417389 & 3.90 & 0.33 & 128.989380 & -2.11 & 0.68 \\
        AGFN-1000 & 32.677258 & 13.40 & 0.14 & 70.010994 & 4.79 & 0.33 & 129.627396 & -1.63 & 0.69 \\
        HBG & 30.82 & 6.94 & 0.15 & 69.90 & 4.62 & 0.36 & 131.82 & 0.04 & 0.84 \\
        \midrule
        OD-Gear-200 & 30.10 & 4.45 & 0.13 & 68.33 & 2.27 & 0.33 & 129.57 & -1.67 & 0.68 \\
        OD-Gear-500 & 31.96 & 10.90 & 0.14 & 67.92 & 1.65 & 0.33 & 126.79 & -3.78 & 0.69 \\
        OD-Gear-1000 & 35.22 & 22.21 & 0.14 & 69.15 & 3.50 & 0.33 & 127.46 & -3.27 & 0.68 \\
        GFACS-noLS & 35.17 & 22.06 & 0.16 & 78.28 & 17.16 & 0.36 & 148.06 & 12.36 & 0.81 \\
        \bottomrule \bottomrule
    \end{tabular}}
\caption{Overall performance comparison on synthetic CVRP instances.  The optimality gap is relative to LKH-3(100), where negative values indicate solutions  cheaper than the reference. OD-Gear rows report ten-run means evaluated without inference-time local search.}
\label{tab:merged-cvrp-synthetic}
\end{table*}

\begin{table*}[!tp]
\centering
\resizebox{\textwidth}{!}{%
\begin{tabular}{lcccccc}
        \toprule \toprule
        Dataset metric & Optimal & AGFN-200 & HBG & GLOP-1000 & OD-Gear-200 & GFACS-noLS \\
        \midrule
        CVRPLib Obj. & 120120.03 & 201463.20 & 171100.72 & 234860.10 & 143520.14 & 231970.83 \\
        CVRPLib Time (s) & - & 43.43 & 27.96 & 70.85 & 44.58 & 16.79 \\
        CVRPLib Gap (\%) & - & 67.72 & 42.44 & 95.52 & 19.48 & 93.12 \\
        TSPLib Obj. & 769.953827 & 915.542687 & 942.17 & 788.482424 & 836.18 & 1464.25 \\
        TSPLib Time (s) & - & 42.90 & 153.46 & 680.62 & 99.85 & 7.69 \\
        TSPLib Gap (\%) & - & 18.91 & 22.37 & 2.41 & 8.60 & 90.17 \\
        \bottomrule \bottomrule
    \end{tabular}}
\caption{Overall performance comparison on CVRPLib and TSPLib. Objectives and times are totals over all instances; gaps are relative to the best-known totals.}
\label{tab:merged-lib}
\end{table*}

\begin{table*}[!tp]
\centering
\resizebox{\textwidth}{!}{%
\begin{tabular}{l|ccc|ccc|ccc}
        \toprule \toprule
        Method & \multicolumn{3}{c|}{$|V|=2000$} & \multicolumn{3}{c|}{$|V|=5000$} & \multicolumn{3}{c}{$|V|=10,000$} \\
        ~ & Obj. & Gap (\%) & Time (s) & Obj. & Gap (\%) & Time (s) & Obj. & Gap (\%) & Time (s) \\
        \midrule
        LKH-3(1000) & 243.569497 & - & 28.49 & 631.602489 & - & 47.00 & 1292.649936 & - & 65.85 \\
        AGFN-200 & 254.051392 & - & 7.55 & 587.557190 & - & 19.53 & 1145.796509 & - & 38.36 \\
        GFACS-200 & 236.277512 & - & 5.56 & 563.053345 & - & 23.33 & 1106.582520 & - & 42.42 \\
        GLOP & 287.393829 & - & 1.01 & 712.800415 & - & 1.76 & 1319.468384 & - & 3.94 \\
        POMO-100 & 648.761200 & - & 3.35 & 2146.482600 & - & 76.47 & - & - & - \\
        HBG & 250.62 & - & 1.71 & 603.99 & - & 5.42 & 1166.56 & - & 11.58 \\
        \midrule
        OD-Gear-200 & 249.93 & - & 1.46 & 582.39 & - & 4.88 & 1210.10 & - & 9.74 \\
        GFACS-noLS & 284.33 & - & 1.50 & 682.98 & - & 4.78 & 1349.50 & - & 10.96 \\
        \bottomrule \bottomrule
    \end{tabular}}
\caption{Ultra-large-scale synthetic CVRP. All learned solvers reuse their small-scale checkpoints; Time is seconds per instance.}
\label{tab:merged-large-cvrp}
\end{table*}

\begin{table*}[!tp]
\centering
\resizebox{\textwidth}{!}{%
\begin{tabular}{l|ccc|ccc|ccc}
        \toprule \toprule
        Method & \multicolumn{3}{c|}{$|V|=200$} & \multicolumn{3}{c|}{$|V|=500$} & \multicolumn{3}{c}{$|V|=1000$} \\
        ~ & Obj. & Gap (\%) & Time (s) & Obj. & Gap (\%) & Time (s) & Obj. & Gap (\%) & Time (s) \\
        \midrule
        OD-Gear-200 & 30.10 & 4.45 & 0.13 & 68.33 & 2.27 & 0.33 & 129.57 & -1.67 & 0.68 \\
        \;\;w/o Decomposition & 32.215363 & 11.79 & 0.14 & 70.047666 & 4.84 & 0.33 & 130.063773 & -1.30 & 0.68 \\
        \;\;w/ Transformer & 33.383432 & 15.85 & 0.14 & 72.551692 & 8.59 & 0.33 & 133.965306 & 1.66 & 0.66 \\
        \midrule
        OD-Gear-500 & 31.96 & 10.90 & 0.14 & 67.92 & 1.65 & 0.33 & 126.79 & -3.78 & 0.69 \\
        \;\;w/o Decomposition & 32.346132 & 12.25 & 0.14 & 69.878954 & 4.59 & 0.33 & 129.427481 & -1.78 & 0.68 \\
        \;\;w/ Transformer & 33.812407 & 17.34 & 0.14 & 73.245449 & 9.63 & 0.32 & 135.287025 & 2.67 & 0.67 \\
        \midrule
        OD-Gear-1000 & 35.22 & 22.21 & 0.14 & 69.15 & 3.50 & 0.33 & 127.46 & -3.27 & 0.68 \\
        \;\;w/o Decomposition & 32.520523 & 12.85 & 0.14 & 70.026472 & 4.81 & 0.33 & 129.541826 & -1.69 & 0.68 \\
        \;\;w/ Transformer & 33.489404 & 16.21 & 0.13 & 74.113616 & 10.93 & 0.32 & 137.635210 & 4.45 & 0.66 \\
        \bottomrule \bottomrule
    \end{tabular}}
\caption{Controlled OD-Gear component ablations on synthetic CVRP, using the main-paper labels w/o Decomposition and w/ Transformer. All entries are averages over 10 independent runs without inference-time local search.}
\label{tab:merged-ablation-odgear}
\end{table*}

\begin{table*}[!tp]
\centering
\resizebox{\textwidth}{!}{%
\begin{tabular}{l|ccc|ccc|ccc}
        \toprule \toprule
        $\hat{N}$ & \multicolumn{3}{c|}{$|V|=200$} & \multicolumn{3}{c|}{$|V|=500$} & \multicolumn{3}{c}{$|V|=1000$} \\
        ~ & Obj. & Gap (\%) & Time (s) & Obj. & Gap (\%) & Time (s) & Obj. & Gap (\%) & Time (s) \\
        \midrule
        10 & 30.592449 & 6.16 & 0.13 & 68.320488 & 2.26 & 0.32 & 128.594177 & -2.41 & 0.64 \\
        30 & 30.519241 & 5.91 & 0.13 & 68.136040 & 1.98 & 0.32 & 128.321060 & -2.62 & 0.64 \\
        50 & 30.509777 & 5.87 & 0.13 & 68.076363 & 1.89 & 0.32 & 128.211121 & -2.70 & 0.63 \\
        70 & 30.457382 & 5.69 & 0.13 & 68.008392 & 1.79 & 0.33 & 128.177887 & -2.73 & 0.66 \\
        100 & 30.426691 & 5.59 & 0.13 & 67.955727 & 1.71 & 0.33 & 128.135544 & -2.76 & 0.66 \\
        110 & 30.424309 & 5.58 & 0.13 & 67.939011 & 1.69 & 0.32 & 128.131516 & -2.76 & 0.66 \\
\bottomrule \bottomrule
    \end{tabular}}
\caption{Sensitivity analysis of the inference group size $\hat{N}$ in OD-Gear-200 w/o Group Sampling on synthetic CVRP (original record).}
\label{tab:merged-hyper-n-oddeal}
\end{table*}

\begin{table*}[!tp]
\centering
\resizebox{\textwidth}{!}{%
\begin{tabular}{l|ccc|ccc|ccc}
        \toprule \toprule
        $\hat{N}$ & \multicolumn{3}{c|}{$|V|=200$} & \multicolumn{3}{c|}{$|V|=500$} & \multicolumn{3}{c}{$|V|=1000$} \\
        ~ & Obj. & Gap (\%) & Time (s) & Obj. & Gap (\%) & Time (s) & Obj. & Gap (\%) & Time (s) \\
        \midrule
        n10 & 30.28 & 5.08 & 0.13 & 68.73 & 2.87 & 0.33 & 130.04 & -1.32 & 0.66 \\
        n30 & 30.18 & 4.72 & 0.13 & 68.51 & 2.54 & 0.33 & 129.79 & -1.50 & 0.66 \\
        n50 & 30.13 & 4.57 & 0.14 & 68.44 & 2.44 & 0.33 & 129.71 & -1.57 & 0.67 \\
        n70 & 30.11 & 4.49 & 0.14 & 68.38 & 2.35 & 0.33 & 129.63 & -1.63 & 0.67 \\
        n100 & 30.10 & 4.45 & 0.14 & 68.34 & 2.29 & 0.33 & 129.57 & -1.67 & 0.68 \\
        n110 & 30.09 & 4.41 & 0.14 & 68.33 & 2.27 & 0.33 & 129.56 & -1.68 & 0.69 \\
\bottomrule \bottomrule
    \end{tabular}}
\caption{Sensitivity analysis of the inference group size $\hat{N}$ ($G_s$) in OD-Gear-200 on synthetic CVRP (ten-run means).}
\label{tab:merged-hyper-n-grpo}
\end{table*}

\begin{table*}[!tp]
\centering
\resizebox{\textwidth}{!}{%
\begin{tabular}{l|ccc|ccc|ccc}
        \toprule \toprule
        $K$ & \multicolumn{3}{c|}{$|V|=200$} & \multicolumn{3}{c|}{$|V|=500$} & \multicolumn{3}{c}{$|V|=1000$} \\
        ~ & Obj. & Gap (\%) & Time (s) & Obj. & Gap (\%) & Time (s) & Obj. & Gap (\%) & Time (s) \\
        \midrule
        20 & 31.350094 & 8.79 & 0.14 & 71.638000 & 7.22 & 0.34 & 141.685013 & 7.52 & 0.68 \\
        30 & 30.533363 & 5.96 & 0.13 & 68.268585 & 2.18 & 0.32 & 132.290024 & 0.39 & 0.67 \\
        40 & 30.426691 & 5.59 & 0.13 & 67.482193 & 1.00 & 0.32 & 128.752258 & -2.29 & 0.65 \\
        50 & 30.450418 & 5.67 & 0.13 & 67.326096 & 0.77 & 0.32 & 127.559982 & -3.20 & 0.65 \\
        60 & 30.564724 & 6.07 & 0.14 & 67.400139 & 0.88 & 0.32 & 127.347626 & -3.36 & 0.65 \\
        \bottomrule \bottomrule
    \end{tabular}}
\caption{Sensitivity analysis of the sparsification degree $K$ in OD-Gear-200 w/o Group Sampling on synthetic CVRP (original record).}
\label{tab:merged-hyper-k-oddeal}
\end{table*}

\begin{table*}[!tp]
\centering
\resizebox{\textwidth}{!}{%
\begin{tabular}{l|ccc|ccc|ccc}
        \toprule \toprule
        $K$ & \multicolumn{3}{c|}{$|V|=200$} & \multicolumn{3}{c|}{$|V|=500$} & \multicolumn{3}{c}{$|V|=1000$} \\
        ~ & Obj. & Gap (\%) & Time (s) & Obj. & Gap (\%) & Time (s) & Obj. & Gap (\%) & Time (s) \\
        \midrule
        k20 & 31.19 & 8.25 & 0.14 & 71.87 & 7.57 & 0.33 & 148.62 & 12.78 & 0.68 \\
        k30 & 30.32 & 5.22 & 0.13 & 67.82 & 1.51 & 0.33 & 133.74 & 1.50 & 0.66 \\
        k40 & 30.22 & 4.88 & 0.13 & 67.14 & 0.48 & 0.32 & 132.27 & 0.38 & 0.67 \\
        k50 & 30.29 & 5.11 & 0.13 & 67.21 & 0.60 & 0.32 & 135.08 & 2.51 & 0.67 \\
        k60 & 30.37 & 5.39 & 0.13 & 67.29 & 0.71 & 0.32 & 134.35 & 1.96 & 0.66 \\
        \bottomrule \bottomrule
    \end{tabular}}
\caption{Sensitivity analysis of the sparsification degree $K$ in OD-Gear-200 on synthetic CVRP.}
\label{tab:merged-hyper-k-grpo}
\end{table*}

\begin{table*}[!tp]
\centering
\begingroup\setlength{\tabcolsep}{2pt}%
\resizebox{\textwidth}{!}{%
\begin{tabular}{l|cc|cccc|cccc|cccc|cccc|cccc}
        \toprule \toprule
        \multicolumn{1}{c|}{Instance} & \multicolumn{2}{c|}{Optimal} & \multicolumn{4}{c|}{AGFN-200} & \multicolumn{4}{c|}{HBG} & \multicolumn{4}{c|}{GLOP} & \multicolumn{4}{c|}{OD-Gear-200} & \multicolumn{4}{c}{GFACS-noLS} \\
        ~ & Obj & Gap & Best & Mean & Gap & Time & Best & Mean & Gap & Time & Best & Mean & Gap & Time & Best & Mean & Gap & Time & Best & Mean & Gap & Time \\
        \midrule
        A-n32-k5 & 784.00 & - & 3174.00 & 3332.20 & 304.85 & 0.44 & 1164.00 & 1168.10 & 48.47 & 0.26 & 1129.00 & 1225.90 & 44.01 & 0.71 & 839.00 & 908.80 & 7.02 & 0.43 & 1275.00 & 1352.20 & 62.63 & 0.15 \\
        A-n33-k5 & 661.00 & - & 765.00 & 787.30 & 15.73 & 0.42 & 749.00 & 758.20 & 13.31 & 0.26 & 1012.00 & 1116.90 & 53.10 & 0.71 & 802.00 & 883.30 & 21.33 & 0.43 & 977.00 & 1055.30 & 47.81 & 0.15 \\
        A-n33-k6 & 742.00 & - & 861.00 & 862.80 & 16.04 & 0.41 & 931.00 & 951.70 & 25.47 & 0.26 & 1067.00 & 1166.40 & 43.80 & 0.71 & 821.00 & 874.30 & 10.65 & 0.44 & 962.00 & 1027.60 & 29.65 & 0.15 \\
        A-n34-k5 & 778.00 & - & 914.00 & 981.30 & 17.48 & 0.42 & 890.00 & 890.00 & 14.40 & 0.26 & 1130.00 & 1184.70 & 45.24 & 0.71 & 919.00 & 951.00 & 18.12 & 0.44 & 1116.00 & 1213.70 & 43.44 & 0.15 \\
        A-n36-k5 & 799.00 & - & 3175.00 & 3325.90 & 297.37 & 0.44 & 977.00 & 1039.50 & 22.28 & 0.26 & 1176.00 & 1231.80 & 47.18 & 0.71 & 1117.00 & 1182.40 & 39.80 & 0.44 & 1259.00 & 1379.10 & 57.57 & 0.15 \\
        A-n37-k5 & 669.00 & - & 744.00 & 785.30 & 11.21 & 0.42 & 742.00 & 747.60 & 10.91 & 0.27 & 1046.00 & 1114.50 & 56.35 & 0.71 & 770.00 & 807.80 & 15.10 & 0.44 & 1038.00 & 1193.10 & 55.16 & 0.16 \\
\bottomrule \bottomrule
    \end{tabular}}
\endgroup
\caption{Per-instance results on the A CVRPLib instances. The results for  AGFN-200, HBG, and GLOP are derived from 10-run evaluations in the original record, whereas OD-Gear-200 and GFACS-noLS report metrics using the 10-run instance benchmark sources. Gaps are computed relative to the best-known solutions.}
\label{tab:cvrplib-A-merged}
\end{table*}

\begin{table*}[!tp]
\centering
\begingroup\setlength{\tabcolsep}{2pt}%
\resizebox{\textwidth}{!}{%
\begin{tabular}{l|cc|cccc|cccc|cccc|cccc|cccc}
        \toprule \toprule
        \multicolumn{1}{c|}{Instance} & \multicolumn{2}{c|}{Optimal} & \multicolumn{4}{c|}{AGFN-200} & \multicolumn{4}{c|}{HBG} & \multicolumn{4}{c|}{GLOP} & \multicolumn{4}{c|}{OD-Gear-200} & \multicolumn{4}{c}{GFACS-noLS} \\
        ~ & Obj & Gap & Best & Mean & Gap & Time & Best & Mean & Gap & Time & Best & Mean & Gap & Time & Best & Mean & Gap & Time & Best & Mean & Gap & Time \\
        \midrule
        E-n101-k14 & 1067.00 & - & 1289.00 & 1317.50 & 20.81 & 0.47 & 1215.00 & 1243.60 & 13.87 & 0.31 & 2096.00 & 2161.20 & 96.44 & 0.74 & 1159.00 & 1182.00 & 8.62 & 0.48 & 1722.00 & 1781.30 & 61.39 & 0.21 \\
        E-n101-k8 & 815.00 & - & 1052.00 & 1073.90 & 29.08 & 0.47 & 957.00 & 968.60 & 17.42 & 0.31 & 1614.00 & 1684.40 & 98.04 & 0.74 & 914.00 & 942.40 & 12.15 & 0.48 & 1131.00 & 1169.60 & 38.77 & 0.20 \\
        E-n22-k4 & 375.00 & - & 436.00 & 457.00 & 16.27 & 0.40 & 449.00 & 454.40 & 19.73 & 0.26 & 521.00 & 565.30 & 38.93 & 0.71 & 419.00 & 440.50 & 11.73 & 0.42 & 542.00 & 600.90 & 44.53 & 0.14 \\
        E-n23-k3 & 569.00 & - & 1414.00 & 1580.40 & 148.51 & 0.42 & 1116.00 & 1144.90 & 96.13 & 0.26 & 675.00 & 769.80 & 18.63 & 0.71 & 867.00 & 907.50 & 52.37 & 0.42 & 1016.00 & 1099.70 & 78.56 & 0.14 \\
        E-n30-k3 & 534.00 & - & 634.00 & 711.90 & 18.73 & 0.42 & 589.00 & 598.70 & 10.30 & 0.26 & 680.00 & 755.30 & 27.34 & 0.71 & 646.00 & 780.40 & 20.97 & 0.43 & 828.00 & 882.60 & 55.06 & 0.15 \\
        E-n33-k4 & 835.00 & - & 4768.00 & 4952.90 & 471.02 & 0.43 & 4228.00 & 4228.60 & 406.35 & 0.28 & 987.00 & 1063.00 & 18.20 & 0.71 & 988.00 & 1058.00 & 18.32 & 0.44 & 1133.00 & 1281.80 & 35.69 & 0.15 \\
\bottomrule \bottomrule
    \end{tabular}}
\endgroup
\caption{Per-instance results on the E CVRPLib instances. The results for AGFN-200, HBG, and GLOP are derived from 10-run evaluations in the original record, whereas OD-Gear-200 and GFACS-noLS report metrics using the 10-run instance benchmark sources. Gaps are computed relative to the best-known solutions.}
\label{tab:cvrplib-E-merged}
\end{table*}

\begin{table*}[!tp]
\centering
\begingroup\setlength{\tabcolsep}{2pt}%
\resizebox{\textwidth}{!}{%
\begin{tabular}{l|cc|cccc|cccc|cccc|cccc|cccc}
        \toprule \toprule
        \multicolumn{1}{c|}{Instance} & \multicolumn{2}{c|}{Optimal} & \multicolumn{4}{c|}{AGFN-200} & \multicolumn{4}{c|}{HBG} & \multicolumn{4}{c|}{GLOP} & \multicolumn{4}{c|}{OD-Gear-200} & \multicolumn{4}{c}{GFACS-noLS} \\
        ~ & Obj & Gap & Best & Mean & Gap & Time & Best & Mean & Gap & Time & Best & Mean & Gap & Time & Best & Mean & Gap & Time & Best & Mean & Gap & Time \\
        \midrule
        B-n31-k5 & 672.00 & - & 3229.00 & 3385.00 & 380.51 & 0.43 & 2559.00 & 2625.90 & 280.80 & 0.27 & 931.00 & 986.30 & 38.54 & 0.71 & 713.00 & 764.20 & 6.10 & 0.43 & 762.00 & 865.50 & 13.39 & 0.15 \\
        B-n34-k5 & 788.00 & - & 3581.00 & 3848.50 & 354.44 & 0.44 & 4040.00 & 4040.00 & 412.69 & 0.28 & 1002.00 & 1100.40 & 27.16 & 0.71 & 959.00 & 1052.70 & 21.70 & 0.44 & 1226.00 & 1344.20 & 55.58 & 0.15 \\
        B-n35-k5 & 955.00 & - & 4415.00 & 4753.10 & 362.30 & 0.44 & 4376.00 & 4483.30 & 358.22 & 0.28 & 1208.00 & 1383.90 & 26.49 & 0.71 & 1175.00 & 1312.60 & 23.04 & 0.44 & 1465.00 & 1554.70 & 53.40 & 0.15 \\
        B-n38-k6 & 805.00 & - & 873.00 & 882.50 & 8.45 & 0.42 & 905.00 & 912.60 & 12.42 & 0.27 & 1137.00 & 1240.20 & 41.24 & 0.71 & 864.00 & 937.30 & 7.33 & 0.45 & 1093.00 & 1150.60 & 35.78 & 0.15 \\
        B-n39-k5 & 549.00 & - & 870.00 & 987.40 & 58.47 & 0.42 & 633.00 & 638.80 & 15.30 & 0.26 & 941.00 & 1073.40 & 71.40 & 0.71 & 692.00 & 726.70 & 26.05 & 0.44 & 831.00 & 897.00 & 51.37 & 0.16 \\
        B-n41-k6 & 829.00 & - & 1365.00 & 1542.90 & 64.66 & 0.43 & 900.00 & 914.60 & 8.56 & 0.27 & 1349.00 & 1419.80 & 62.73 & 0.71 & 936.00 & 975.10 & 12.91 & 0.45 & 991.00 & 1065.40 & 19.54 & 0.16 \\
\bottomrule \bottomrule
    \end{tabular}}
\endgroup
\caption{Per-instance results on the B CVRPLib instances. The results for AGFN-200, HBG, and GLOP are derived from 10-run evaluations in the original record, whereas OD-Gear-200 and GFACS-noLS report metrics using the 10-run instance benchmark sources. Gaps are computed relative to the best-known solutions.}
\label{tab:cvrplib-B-merged}
\end{table*}

\begin{table*}[!tp]
\centering
\begingroup\setlength{\tabcolsep}{2pt}%
\resizebox{\textwidth}{!}{%
\begin{tabular}{l|cc|cccc|cccc|cccc|cccc|cccc}
        \toprule \toprule
        \multicolumn{1}{c|}{Instance} & \multicolumn{2}{c|}{Optimal} & \multicolumn{4}{c|}{AGFN-200} & \multicolumn{4}{c|}{HBG} & \multicolumn{4}{c|}{GLOP} & \multicolumn{4}{c|}{OD-Gear-200} & \multicolumn{4}{c}{GFACS-noLS} \\
        ~ & Obj & Gap & Best & Mean & Gap & Time & Best & Mean & Gap & Time & Best & Mean & Gap & Time & Best & Mean & Gap & Time & Best & Mean & Gap & Time \\
        \midrule
        tai100a & 2041.34 & - & 2541.00 & 2777.40 & 24.48 & 0.47 & 2588.56 & 2642.83 & 26.81 & 0.32 & 3938.00 & 4065.70 & 92.91 & 0.75 & 2146.33 & 2169.38 & 5.14 & 0.48 & 4540.34 & 4753.11 & 122.42 & 0.20 \\
        tai100b & 1939.90 & - & 2827.00 & 2870.60 & 45.73 & 0.48 & 2652.75 & 2684.16 & 36.75 & 0.32 & 3661.00 & 3887.00 & 88.72 & 0.74 & 2165.94 & 2290.41 & 11.65 & 0.48 & 4382.38 & 4500.78 & 125.91 & 0.20 \\
        tai100c & 1406.20 & - & 1889.00 & 1914.90 & 34.33 & 0.48 & 1918.29 & 1969.51 & 36.42 & 0.32 & 3174.00 & 3423.20 & 125.71 & 0.74 & 1692.30 & 1745.56 & 20.35 & 0.48 & 5377.96 & 5660.88 & 282.45 & 0.19 \\
        tai100d & 1580.46 & - & 2639.00 & 2698.20 & 66.98 & 0.48 & 2105.92 & 2141.14 & 33.25 & 0.32 & 2995.00 & 3165.80 & 89.50 & 0.75 & 1768.79 & 1800.94 & 11.92 & 0.48 & 2849.61 & 3012.96 & 80.30 & 0.20 \\
        tai150a & 3055.23 & - & 3991.00 & 4236.70 & 30.63 & 0.50 & 3782.93 & 3818.93 & 23.82 & 0.34 & 6600.00 & 6802.30 & 116.02 & 0.76 & 3402.00 & 3459.28 & 11.35 & 0.50 & 6851.34 & 7312.41 & 124.25 & 0.23 \\
        tai150b & 2727.03 & - & 4535.00 & 4702.80 & 66.30 & 0.51 & 4039.19 & 4117.51 & 48.12 & 0.36 & 5954.00 & 6577.70 & 118.33 & 0.76 & 2962.11 & 2987.97 & 8.62 & 0.51 & 5614.34 & 6028.37 & 105.88 & 0.23 \\
\bottomrule \bottomrule
    \end{tabular}}
\endgroup
\caption{Per-instance results on the Tai CVRPLib instances. The results for AGFN-200, HBG, and GLOP are derived from 10-run evaluations in the original record,  whereas OD-Gear-200 and GFACS-noLS report metrics using the 10-run instance benchmark sources. Gaps are computed relative to the best-known solutions.}
\label{tab:cvrplib-tai-merged}
\end{table*}

\begin{table*}[!tp]
\centering
\begingroup\setlength{\tabcolsep}{2pt}%
\resizebox{\textwidth}{!}{%
\begin{tabular}{l|cc|cccc|cccc|cccc|cccc|cccc}
        \toprule \toprule
        \multicolumn{1}{c|}{Instance} & \multicolumn{2}{c|}{Optimal} & \multicolumn{4}{c|}{AGFN-200} & \multicolumn{4}{c|}{HBG} & \multicolumn{4}{c|}{GLOP} & \multicolumn{4}{c|}{OD-Gear-200} & \multicolumn{4}{c}{GFACS-noLS} \\
        ~ & Obj & Gap & Best & Mean & Gap & Time & Best & Mean & Gap & Time & Best & Mean & Gap & Time & Best & Mean & Gap & Time & Best & Mean & Gap & Time \\
        \midrule
        P-n101-k4 & 681.00 & - & 893.00 & 922.40 & 31.13 & 0.47 & 838.00 & 874.10 & 23.05 & 0.31 & 1169.00 & 1230.00 & 71.66 & 0.74 & 792.00 & 801.90 & 16.30 & 0.47 & 913.00 & 970.50 & 34.07 & 0.19 \\
        P-n16-k8 & 450.00 & - & 645.00 & 723.80 & 43.33 & 0.41 & 760.00 & 760.00 & 68.89 & 0.26 & 472.00 & 502.30 & 4.89 & 0.71 & 479.00 & 516.80 & 6.44 & 0.43 & 473.00 & 490.40 & 5.11 & 0.14 \\
        P-n19-k2 & 212.00 & - & 809.00 & 912.30 & 281.60 & 0.41 & 883.00 & 883.00 & 316.51 & 0.26 & 259.00 & 288.20 & 22.17 & 0.70 & 328.00 & 411.80 & 54.72 & 0.42 & 405.00 & 424.40 & 91.04 & 0.14 \\
        P-n20-k2 & 216.00 & - & 891.00 & 980.70 & 312.50 & 0.42 & 572.00 & 572.00 & 164.81 & 0.26 & 247.00 & 290.20 & 14.35 & 0.70 & 381.00 & 450.10 & 76.39 & 0.42 & 394.00 & 430.60 & 82.41 & 0.14 \\
        P-n21-k2 & 211.00 & - & 924.00 & 1008.00 & 337.91 & 0.42 & 562.00 & 590.40 & 166.35 & 0.26 & 261.00 & 293.40 & 23.70 & 0.70 & 349.00 & 418.90 & 65.40 & 0.42 & 398.00 & 425.00 & 88.63 & 0.14 \\
        P-n22-k2 & 216.00 & - & 949.00 & 1042.90 & 339.35 & 0.42 & 628.00 & 660.70 & 190.74 & 0.26 & 239.00 & 281.50 & 10.65 & 0.70 & 356.00 & 439.50 & 64.81 & 0.41 & 414.00 & 443.90 & 91.67 & 0.14 \\
\bottomrule \bottomrule
    \end{tabular}}
\endgroup
\caption{Per-instance results on the P CVRPLib instances. The results for  AGFN-200, HBG, and GLOP are derived from 10-run evaluations in the original record; OD-Gear-200 and GFACS-noLS columns use the 10-run instance benchmark sources. Gaps are computed  relative to the best-known solutions.}
\label{tab:cvrplib-P-merged}
\end{table*}

\begin{table*}[!tp]
\centering
\begingroup\setlength{\tabcolsep}{2pt}%
\resizebox{\textwidth}{!}{%
\begin{tabular}{l|cc|ccc|ccc|ccc}
        \toprule \toprule
        Instance & \multicolumn{2}{c|}{AGFN-200} & \multicolumn{3}{c|}{HBG} & \multicolumn{3}{c|}{OD-Gear-200} & \multicolumn{3}{c}{GFACS-noLS} \\
        ~ & Best & Mean & Best & Mean & Time & Best & Mean & Time & Best & Mean & Time \\
        \midrule
        XL-n10001-k1570 & 2418.194 & 2419.821 & 2458.60 & 2461.59 & 12.35 & 2527.03 & 2529.42 & 11.88 & 3210.53 & 3219.86 & 8.49 \\
        XL-n1048-k237 & 575.444 & 577.932 & 794.86 & 800.20 & 1.45 & 415.12 & 416.06 & 1.30 & 793.29 & 803.45 & 0.92 \\
        XL-n1094-k157 & 132.149 & 132.149 & 120.03 & 120.42 & 1.22 & 127.46 & 127.71 & 1.27 & 139.72 & 141.02 & 0.90 \\
        XL-n1141-k112 & 112.058 & 112.058 & 109.18 & 109.75 & 1.24 & 120.13 & 120.43 & 1.27 & 124.06 & 126.54 & 0.90 \\
        XL-n1188-k96 & 116.018 & 116.018 & 118.63 & 119.40 & 1.26 & 126.76 & 128.35 & 1.29 & 142.14 & 143.40 & 0.91 \\
        XL-n1234-k55 & 107.630 & 107.630 & 112.86 & 113.52 & 1.28 & 133.50 & 133.88 & 1.29 & 145.58 & 149.07 & 0.90 \\
\bottomrule \bottomrule
    \end{tabular}}
\endgroup
\caption{Per-instance results on the XL CVRP benchmark (part 1 of 3). The AGFN-200 columns reproduce the original repository values from its official repository under the paper's scale.  The objective values of HBG, OD-Gear, and GFACS-noLS are divided by 1,000 to match that scale. Time is seconds per instance.}
\label{tab:cvrp-xl-merged-1}
\end{table*}

\begin{table*}[!tp]
\centering
\begingroup\setlength{\tabcolsep}{2pt}%
\resizebox{\textwidth}{!}{%
\begin{tabular}{l|cc|ccc|ccc|ccc}
        \toprule \toprule
        Instance & \multicolumn{2}{c|}{AGFN-200} & \multicolumn{3}{c|}{HBG} & \multicolumn{3}{c|}{OD-Gear-200} & \multicolumn{3}{c}{GFACS-noLS} \\
        ~ & Best & Mean & Best & Mean & Time & Best & Mean & Time & Best & Mean & Time \\
        \midrule
        XL-n2587-k66 & 103.948 & 103.948 & 93.63 & 94.55 & 2.47 & 128.66 & 129.18 & 2.48 & 169.31 & 173.94 & 1.83 \\
        XL-n2634-k17 & 115.069 & 115.071 & 48.80 & 49.30 & 2.52 & 51.85 & 52.16 & 2.47 & 107.94 & 111.42 & 1.81 \\
        XL-n2681-k540 & 847.197 & 847.197 & 886.89 & 888.05 & 2.88 & 825.92 & 826.41 & 2.94 & 1677.38 & 1689.08 & 2.10 \\
        XL-n2727-k546 & 437.412 & 437.441 & 450.43 & 450.77 & 2.95 & 451.99 & 452.88 & 3.02 & 1443.36 & 1447.52 & 2.14 \\
        XL-n2774-k286 & 430.153 & 430.168 & 430.89 & 432.07 & 2.78 & 432.22 & 432.75 & 2.83 & 510.47 & 515.86 & 2.01 \\
        XL-n2821-k208 & 244.114 & 244.114 & 269.49 & 270.26 & 2.79 & 260.59 & 261.02 & 2.82 & 676.87 & 682.06 & 1.98 \\
\bottomrule \bottomrule
    \end{tabular}}
\endgroup
\caption{Per-instance results on the XL CVRP benchmark (part 2 of 3). The AGFN-200 columns reproduce the original repository values from its official repository under the paper's scale.  The objective values of HBG, OD-Gear, and GFACS-noLS are divided by 1,000 to match that scale. Time is seconds per instance.}
\label{tab:cvrp-xl-merged-2}
\end{table*}

\begin{table*}[!tp]
\centering
\begingroup\setlength{\tabcolsep}{2pt}%
\resizebox{\textwidth}{!}{%
\begin{tabular}{l|cc|ccc|ccc|ccc}
        \toprule \toprule
        Instance & \multicolumn{2}{c|}{AGFN-200} & \multicolumn{3}{c|}{HBG} & \multicolumn{3}{c|}{OD-Gear-200} & \multicolumn{3}{c}{GFACS-noLS} \\
        ~ & Best & Mean & Best & Mean & Time & Best & Mean & Time & Best & Mean & Time \\
        \midrule
        XL-n4951-k203 & 310.547 & 310.854 & 314.37 & 315.60 & 5.27 & 332.43 & 332.84 & 5.38 & 429.20 & 442.12 & 3.57 \\
        XL-n5061-k184 & 230.329 & 230.329 & 213.44 & 215.20 & 5.48 & 189.98 & 190.44 & 5.55 & 304.09 & 310.39 & 3.70 \\
        XL-n5174-k55 & 175.528 & 175.574 & 124.79 & 125.87 & 5.76 & 94.32 & 95.14 & 5.55 & 218.16 & 223.70 & 3.74 \\
        XL-n5288-k1246 & 2046.394 & 2047.613 & 2117.34 & 2119.62 & 6.84 & 2082.47 & 2083.47 & 6.94 & 4032.96 & 4056.87 & 4.56 \\
        XL-n5406-k783 & 1076.376 & 1076.561 & 1090.12 & 1091.04 & 6.52 & 1073.37 & 1074.09 & 6.59 & 1673.30 & 1682.85 & 4.32 \\
        XL-n5526-k553 & 502.272 & 502.925 & 434.88 & 435.95 & 7.20 & 357.98 & 358.29 & 6.57 & 441.88 & 445.36 & 4.36 \\
\bottomrule \bottomrule
    \end{tabular}}
\endgroup
\caption{Per-instance results on the XL CVRP benchmark (part 3 of 3). The AGFN-200 columns reproduce the original repository values from its official repository under the paper's scale. The objective values of HBG, OD-Gear, and GFACS-noLS are divided by 1,000 to match that scale. Time is seconds per instance.}
\label{tab:cvrp-xl-merged-3}
\end{table*}

\begin{table*}[!tp]
\centering
\resizebox{\textwidth}{!}{%
\begin{tabular}{l|ccc|ccc|ccc}
        \toprule \toprule
        Method & \multicolumn{3}{c|}{$|V|=200$} & \multicolumn{3}{c|}{$|V|=500$} & \multicolumn{3}{c}{$|V|=1000$} \\
        ~ & Obj. & Gap (\%) & Time (s) & Obj. & Gap (\%) & Time (s) & Obj. & Gap (\%) & Time (s) \\
        \midrule
        LKH-3(100) & 10.724371 & - & 0.40 & 16.606491 & - & 1.04 & 23.282794 & - & 2.05 \\
        LKH-3(1000) & 10.719438 & -0.05 & 2.11 & 16.562766 & -0.26 & 5.12 & 23.177903 & -0.45 & 11.17 \\
        LKH-3(10,000) & 10.719180 & -0.05 & 10.87 & 16.551830 & -0.33 & 26.16 & 23.143115 & -0.60 & 54.74 \\
        POMO & 10.887800 & 1.52 & 0.05 & 20.181600 & 21.53 & 0.47 & 32.505500 & 39.61 & 3.66 \\
        NeuOpt & 10.837507 & 1.05 & 0.45 & 55.235378 & 232.61 & 1.34 & 277.531128 & 1092.00 & 3.61 \\
        GLOP & 10.856474 & 1.23 & 0.09 & 16.908892 & 1.82 & 0.62 & 23.837715 & 2.38 & 1.25 \\
        ACO & 11.505671 & 7.29 & 3.39 & 17.970035 & 8.21 & 34.89 & 25.137234 & 7.96 & 266.98 \\
        GFACS & 10.801056 & 0.72 & 0.15 & 16.957058 & 2.11 & 0.78 & 23.949020 & 2.86 & 4.32 \\
        AGFN-200 & 11.841615 & 10.42 & 0.09 & 18.568083 & 11.81 & 0.21 & 26.460453 & 13.65 & 0.41 \\
        AGFN-500 & 12.041571 & 12.28 & 0.09 & 18.790956 & 13.15 & 0.21 & 26.460077 & 13.65 & 0.41 \\
        AGFN-1000 & 12.274513 & 14.45 & 0.12 & 19.237740 & 15.84 & 0.28 & 27.059122 & 16.22 & 0.50 \\
        HBG & 11.74 & 9.47 & 0.14 & 18.65 & 12.29 & 0.73 & 26.87 & 15.41 & 2.70 \\
        \midrule
        OD-Gear-200 & 13.36 & 24.54 & 0.06 & 21.57 & 29.89 & 0.13 & 31.96 & 37.25 & 0.26 \\
        OD-Gear-500 & 13.89 & 29.49 & 0.05 & 21.03 & 26.61 & 0.13 & 29.92 & 28.51 & 0.26 \\
        OD-Gear-1000 & 17.70 & 65.04 & 0.05 & 25.24 & 52.02 & 0.13 & 34.23 & 47.03 & 0.27 \\
        GFACS-noLS & 14.62 & 36.34 & 0.02 & 26.00 & 56.55 & 0.02 & 41.97 & 80.24 & 0.03 \\
        \bottomrule \bottomrule
    \end{tabular}}
\caption{Overall performance comparison on synthetic TSP (gaps are relative to LKH-3(100)).}
\label{tab:merged-tsp-synthetic}
\end{table*}

\section{Further Discussion and Limitations}
\label{appendix_discussion}
The OD-Gear framework effectively overcomes the scalability bottlenecks of neural solvers in large-scale CVRP by integrating dynamic expert guidance with an online decomposition mechanism. This approach achieves a superior balance between solution quality and inference efficiency across complex topologies at the 10,000-node scale.

Nonetheless, there remains scope for further optimization. Experimental results indicate that for tasks dominated by pure geometric constraints, such as the traveling salesman problem (TSP), our constructive strategy may be slightly outperformed by specialized algorithms focusing on geometric spatial partitioning. 
This suggests that incorporating inductive biases specifically tailored to simple topologies could further enhance performance. Furthermore, while the current generator implicitly internalizes decomposition logic, integrating explicit clustering-based reward mechanisms into the training objective remains a promising direction for refining the model’s latent space partitioning.

As future work, we aim to expand the applicability of OD-Gear to complex real-world constraints, such as vehicle routing problems with time windows (VRPTW) or pickup and delivery problems (PDP). From an industrial perspective, OD-Gear exhibits significant potential for dynamic logistics scheduling. Its sub-second inference capability is well- suited for dynamic logistics scheduling scenarios that demand ultra-fast response times, such as instant delivery and ride-hailing route planning, offering substantial efficiency gains for real-world supply chain management.

\twocolumn[\begingroup\noindent\hbox to \columnwidth{\hfil{\Large\bfseries\refname}\hfil}\par\vskip 3pt\endgroup]
\makeatletter
\renewcommand{\bibsection}{}
\let\ODGear@orig@bibsetup\@bibsetup
\renewcommand{\@bibsetup}[1]{%
  \ODGear@orig@bibsetup{#1}%
  \setlength{\topsep}{0pt plus 0pt}%
  \setlength{\partopsep}{0pt plus 0pt}%
  \setlength{\itemsep}{0pt plus 0pt}%
  \setlength{\parsep}{0pt plus 0pt}%
}
\makeatother
\setlength{\bibsep}{0pt plus 0pt}
\setlength{\parskip}{0pt plus 0pt}
\bibliography{aaai2027}

\end{document}